%% file: neurips_2026.tex
\documentclass{article}

\usepackage[preprint]{neurips_2026}

\usepackage[utf8]{inputenc} 
\usepackage[T1]{fontenc}    
\usepackage{hyperref}       
\usepackage{url}            
\usepackage{booktabs}       
\usepackage{amsfonts}       
\usepackage{nicefrac}       
\usepackage{microtype}      
\usepackage[table]{xcolor}    
\usepackage{graphicx}        
\usepackage{multirow}
\usepackage{placeins}
\usepackage{algorithm}
\usepackage{algorithmic}
\usepackage{amsmath}        
\usepackage{tcolorbox}
\tcbuselibrary{breakable, skins}
\usepackage{graphicx, caption}
\definecolor{deepblue}{RGB}{70,130,180}

\title{VideoSeeker: Incentivizing Instance-level Video Understanding via Native Agentic Tool Invocation}

\author{Yiming Zhao$^{1,2}\thanks{\quad Equal Contribution}$, Yu Zeng$^{1,2*}$,  Wenxuan Huang$^{2,3*}$, Zhen Fang$^{1,2*}$, Qing Miao$^{4}$, \\[2pt]
\textbf{Qisheng Su$^{1}$, Jiawei Zhao$^{2}$, Jiayin Cai$^{2}$, Lin Chen$^{1}$, Zehui Chen$^{1}$} \\[2pt]
\textbf{Yukun Qi$^{1}$,Yao Hu$^{2}$,Xiaolong Jiang$^{2}$, Feng Zhao$^{1}\thanks{\quad Corresponding Author}$} \\[2pt]
$^{1}$ University of Science and Technology of China \quad
$^{2}$ Xiaohongshu Inc.\\
$^{3}$ East China Normal University \quad
$^{4}$ Xi’an Jiaotong University \\
Project Page: {\url{https://gaotiexinqu.github.io/VideoSeeker/}}
}

\begin{document}

\maketitle

\input{section/0_abstract}
\input{section/1_introduction}
\input{section/2_related_works}
\input{section/3_method}
\input{section/4_experiment}
\input{section/5_conclusion}

\newpage
{\small
    \bibliographystyle{abbrv}
    \bibliography{reference}
}


\newpage
\appendix
\input{section/6_appendix}


\newpage
\input{section/99_check_list}

\end{document}

%% file: section/0_abstract.tex
\begin{abstract}

Large Vision-Language Models (LVLMs) have shown significant progress in video understanding, yet they face substantial challenges in tasks requiring precise spatiotemporal localization at the instance level. Existing methods primarily rely on text prompts for human-model interaction, but these prompts struggle to provide precise spatial and temporal references, resulting in poor user experience. Furthermore, current approaches typically decouple visual perception from language reasoning, centering reasoning around language rather than visual content, which limits the model's ability to proactively perceive fine-grained visual evidence. To address these challenges, we propose \textbf{VideoSeeker}, a novel paradigm for instance-level video understanding through visual prompts. VideoSeeker seamlessly integrates agentic reasoning with instance-level video understanding tasks, enabling the model to proactively perceive and retrieve relevant video segments on demand. We construct a four-stage fully automated data synthesis pipeline to efficiently generate large-scale, high-quality instance-level video data. We internalize tool-calling and proactive perception capabilities into the model via cold-start supervision and RL training, building a powerful video understanding model. Experiments demonstrate that our model achieves an average improvement of +13.7\% over baselines on instance-level video understanding tasks, surpassing powerful closed-source models such as GPT-4o and Gemini-2.5-Pro, while also showing effective transferability on general video understanding benchmarks. The relevant datasets and code will be released publicly.
\end{abstract}

%% file: section/1_introduction.tex
\section{Introduction}

\begin{figure*}[!htp]
    \centering
    \includegraphics[width=0.95\textwidth]{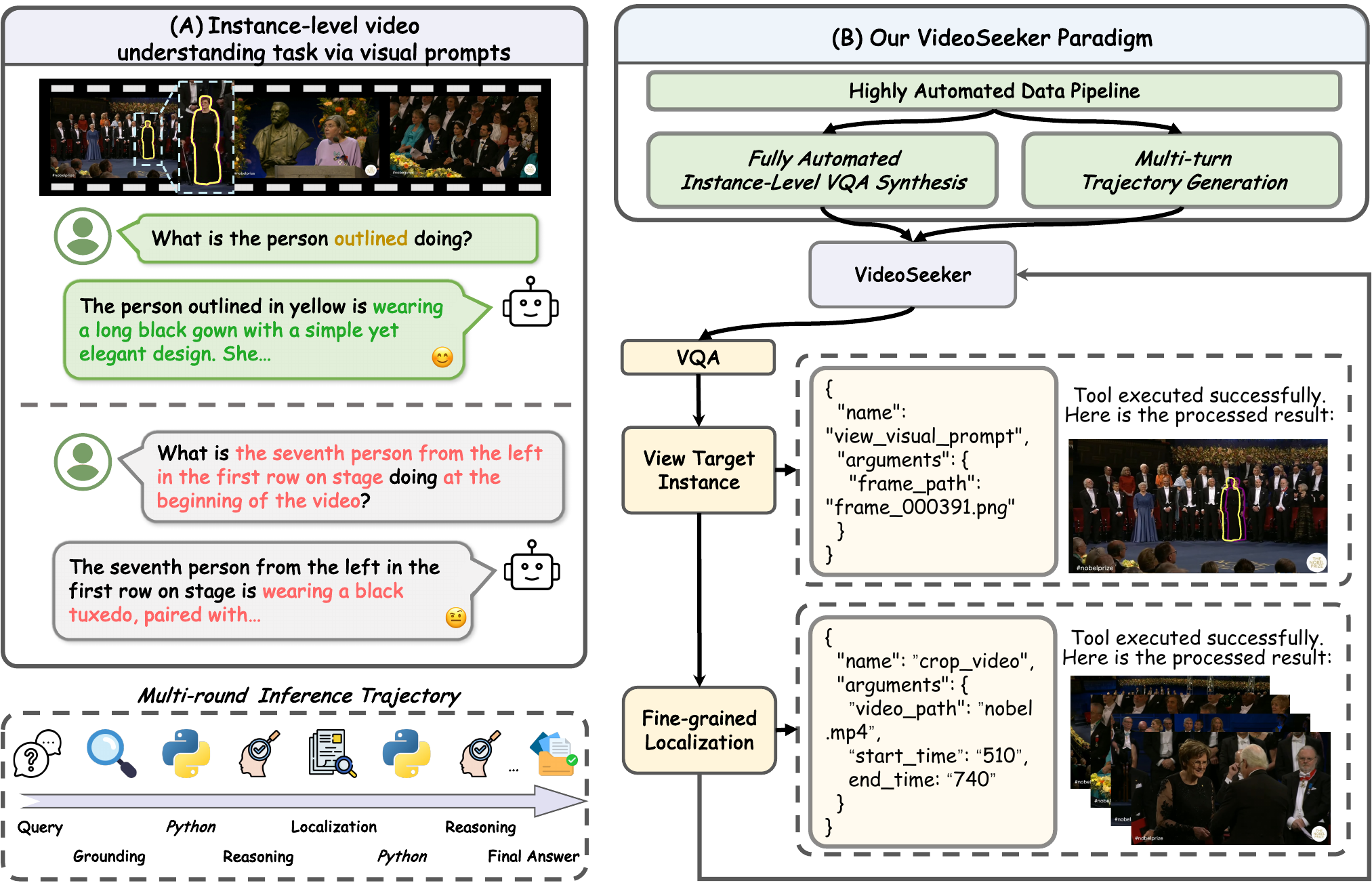}
    \caption{
        \textbf{Overview of VideoSeeker.} \textit{(A): Instance-level video understanding tasks} require models to accurately locate and reason about specific instances in videos guided by visual prompts, given a video, a visual prompt frame, and a query. Compared to text-only prompts that require lengthy referential descriptions, visual prompts provide a more intuitive interaction method. \textit{(B): Pipeline overview.} We design a four-stage pipeline to construct instance-level video data, followed by a two-stage training strategy to integrate multimodal instance-level video understanding capabilities.
    }
    \vspace{-20pt}
    \label{fig:main}
\end{figure*}

Large Vision Language Models (LVLMs) have achieved significant progress in recent years, demonstrating exceptional capabilities across diverse tasks including image captioning \citep{zeng2025caption,deitke2025molmo, xing2025caprl,clark2026molmo2}, visual question answering \citep{chen2024sharegpt4v, bai2025qwen3,zeng2025jigsaw,xu2025llavacot,chen2024mmstar}, video understanding \citep{zhao2025v2pbench,li2024videomme,qi2025vcrbench,hong2026glm5vturbo,wang2025internvl3p5,ren2024timechat}, and complex multimodal reasoning \citep{team2026kimik2p5,chen2025minimax}. By deeply integrating visual and textual modalities, these models have developed strong multimodal perception and reasoning capabilities. Recently, methods \citep{feng2025videor1,wang2025videorft,liu2025videothinker} have successfully introduced reinforcement learning (RL) into video question answering and temporal localization. By leveraging environmental reward signals to guide models in exploring superior reasoning strategies, these approaches have achieved remarkable performance improvements in video understanding tasks, further expanding the temporal reasoning capabilities of LVLMs.

However, existing methods still suffer from two key limitations. \textit{(1) Most current approaches decouple visual perception from language reasoning}, centering reasoning on language rather than visual evidence \citep{feng2025videor1, wang2025videorft, liu2025videothinker}. This weakens visual reasoning and often causes hallucinations in long-video scenarios \citep{yang2025longvt}. Moreover, the widely used single-pass uniform sampling strategy is a passive perception mechanism that cannot adaptively capture key visual evidence, frequently missing fine-grained details critical for reasoning \citep{li2024videomme}. As a result, such methods struggle with precise localization tasks, e.g., identifying when a person appears for the second time. \textit{(2) Existing methods and benchmarks mainly focus on holistic video understanding} \citep{li2024videomme, fu2024longvideobench}, emphasizing global semantics and coarse-grained events while lacking fine-grained spatio-temporal localization and reasoning for specific instances \citep{wang2025timer1}. In addition, current approaches rely solely on text queries (Figure \ref{fig:main}. A), which cannot provide precise spatial-temporal references \citep{zhao2025v2pbench}. This makes evaluating LVLMs in complex multi-object scenarios difficult and forces users to describe targets with lengthy referential language, reducing interaction efficiency and user experience.

\begin{figure*}[t]
    \centering
    \includegraphics[width=0.98\textwidth]{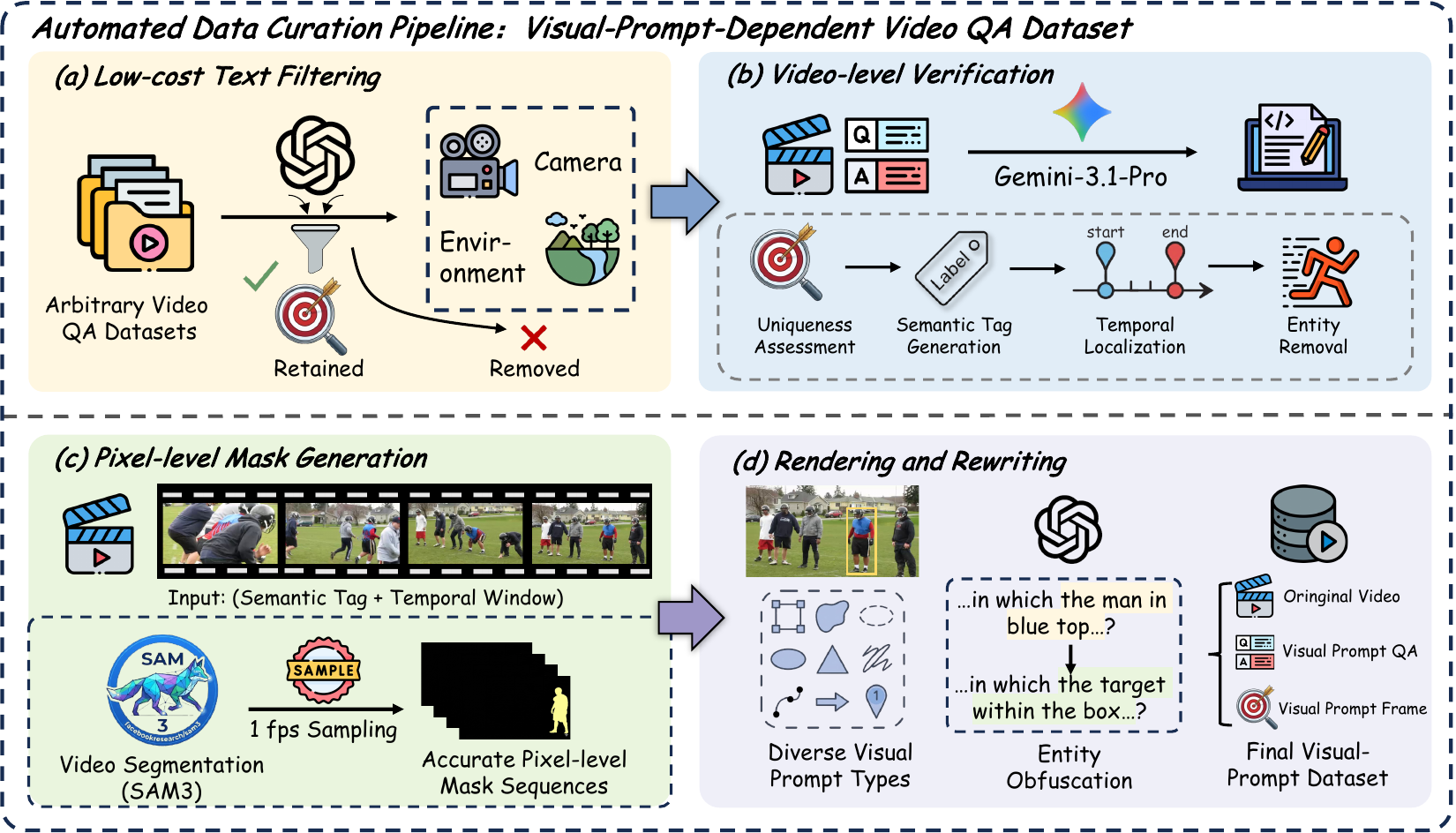}
    \caption{
        \textbf{Our Data Pipeline.} \textit{(1) Low-cost Text Filtering} rapidly filters pure text QA pairs; \textit{(2) Video-level Verification} verifies target uniqueness and generates semantic tags; \textit{(3) Pixel-level Mask Generation} produces pixel-wise masks via SAM3; \textit{(4) Visual Prompt Rendering} renders diverse visual prompt types and rewrites QA to depend on them.
    }
    \vspace{-10pt}
    \label{fig:data_pipeline}
\end{figure*}

To address these issues, we propose \textbf{VideoSeeker}, a novel paradigm for instance-level video understanding based on visual prompts (Figure \ref{fig:main}. B). Unlike text-based prompts that rely on language descriptions, visual prompts enable users to directly annotate target regions on video frames, achieving more precise spatial and temporal references. As illustrated in Figure \ref{fig:data_pipeline}, we construct a four-stage fully automated visual prompt video question answering data synthesis pipeline to obtain high-quality data. Subsequently, through a two-stage strategy of SFT for cold-start combined with Agentic RL, we guide the model to explore the policy space with high information gain, ultimately integrating multi-round agentic reasoning paradigms and instance-level video understanding tasks into the baseline model. In the data pipeline, we first employ a lightweight language model for low-cost text pre-screening, then leverage powerful video understanding models to perform target uniqueness verification ensuring question solvability. Additionally, we integrate SAM3 \cite{sam32025} to achieve pixel-level instance segmentation, ultimately rendering diverse visual prompt types and generating instance-level video QA data ready for training. Extensive experiments demonstrate that our proposed VideoSeeker significantly outperforms all open-source baselines on the instance-level video understanding benchmark V2P-Bench, with our 8B model achieving an average improvement of +13.7\% over baseline, surpassing powerful closed-source models such as GPT-4o and Gemini-2.5-Pro, while also exhibiting effective transferability to general video understanding scenarios.

In a nutshell, our contributions are as follows:

\begin{itemize}
\item We propose VideoSeeker, an agentic instance-level video understanding paradigm. By organically integrating agentic reasoning, VideoSeeker breaks through the limitations of text queries and achieves more precise references.
\item We construct a four-stage instance-level video question answering data synthesis pipeline and efficiently generates large-scale, high-quality instance-level video data, providing an effective solution to the scarcity of relevant training data.
\item Extensive experiments demonstrate that VideoSeeker significantly outperforms all open-source and proprietary baselines on instance-level video understanding tasks, while also exhibiting effective transferability to general video understanding scenarios.
\end{itemize}

%% file: section/2_related_works.tex
\section{Related Works}

\textbf{Reinforcement Learning for Vision Language Models.}
Inspired by the success of large reasoning models such as OpenAI o1~\citep{openai2024o1} and DeepSeek-R1~\citep{guo2025deepseekr1},
recent studies extend GRPO-style RL~\citep{shao2024grpo} from text-only reasoning to multimodal domains~\citep{rafailov2023direct}.
In vision, methods enhance reasoning for image QA~\citep{huang2025visionr1,meng2025mmeureka,deng2025openvlthinker},
grounding~\citep{liu2025visualrft,shen2025vlmr1}. For example, Perception-R1~\citep{yu2025perception} leverages object matching and IoU as reward signals to improve grounding,
and DeepEyes~\citep{zheng2025deepeyes} shows how RL can encourage models to invoke visual tools, thereby expanding perceptual abilities.
Video-centric approaches further tackle temporal reasoning tasks such as video QA~\citep{feng2025videor1,wang2025videorft}
and temporal grounding~\citep{wang2025timer1,li2025videochatr1}, with Video-R1~\citep{feng2025videor1},
VideoChat-R1~\citep{li2025videochatr1} and VideoRFT~\citep{wang2025videorft} being representative works.
Additionally, Vision-R1~\citep{huang2025visionr1} and R1-OneVision~\citep{yang2025r1} construct multimodal CoT datasets
by converting visual information into textual representations to support stronger reasoning.
Despite these advances, most methods still rely on text-based CoT reasoning~\citep{feng2025videor1,li2025videochatr1,chen2025scaling},
which remains largely language-centric~\citep{yang2025longvt}, limiting visual reasoning and increasing hallucinations
in long-video scenarios. This motivates us to explore how to enable more effective video reasoning
through visual tool augmentation.

\textbf{Tool-Augmented Agentic Vision Language Models.}
Recent advances in LVLMs show that equipping models with external tools can enhance capabilities beyond pure text understanding and generation~\citep{su2025pixelreasoner,zheng2025deepeyes}.
In the image domain, methods~\citep{zheng2025deepeyes,su2025pixelreasoner,openai2025o3,wang2025adatooler,hong2025deepeyesv2} enable MLLMs to ``think with images'' by integrating visual tools for image reasoning, while VILA-SR~\citep{wu2025vilasr} reinforces spatial reasoning with interwoven visual drawing. In the video domain, LongVT~\citep{yang2025longvt} proposes iMCoTT that enables MLLMs to perform native temporal retrieval and reasoning by dynamically selecting and re-inspecting relevant video segments, without an auxiliary retriever. VITAL~\citep{zhang2025thinking} constructs a visual toolbox that allows models to densely sample new video frames
on demand during reasoning, enabling precise long video reasoning.
Additionally, Ego-R1~\citep{tian2025egor1} explores chain-of-tool-thought reasoning in first-person videos, and PyVision~\citep{zhao2025pyvision} proposes dynamic tool calling. However, our method differs from prior works such as LongVT~\citep{yang2025longvt} and VITAL~\citep{zhang2025thinking} in the following key aspects: \textit{(1) VideoSeeker targets instance-level video understanding tasks}, focusing on precise localization and tracking of specific target instances within videos; whereas LongVT and VITAL primarily emphasize holistic semantic modeling. \textit{(2) VideoSeeker employs visual prompts (e.g., bounding boxes, points, and masks) as queries}, enabling direct specification of target instances with more precise spatial and temporal references; whereas prior works rely entirely on pure text queries, requiring extensive referential language to describe targets. \textit{(3) We design a four-stage fully automated data pipeline} that efficiently generates large-scale, high-quality instance-level video data, and propose a two-stage training paradigm to internalize native tool-calling capabilities into the base model, enabling native instance-level video understanding.

%% file: section/3_method.tex
\section{Method}

\subsection{Task Formulation And Environmental Interaction}
\label{sec:task_formulation}

\textbf{Task Formulation.} Given a query \( Q \), a visual prompt frame \(\mathcal{F}_{vp}\) and a video \(\mathcal{V}\) of arbitrary length, the goal of instance-level video understanding is to accurately answer the query \( Q \) with respect to the specific instance indicated by \(\mathcal{F}_{vp}\), and output a grounded answer \( A \). Unlike general video question answering where the answer is independent of a particular object, instance-level video understanding requires the model to \textit{(1)} precisely associate the visual prompt with the corresponding target instance in \(\mathcal{V}\) and \textit{(2)} reason about the temporal dynamics of that specific instance across \(\mathcal{V}\) to produce the final answer \( A \).

\textbf{Environmental Interaction.} The policy model \(\pi_{\theta}\) interacts with the video environment through multi-turn active perception control, rather than passively encoding all context in a single pass. Specifically, the model is equipped with a perception tool set \(\mathcal{T} = \{view\_visual\_prompt,\ crop\_video\}\): the former continuously provides visual prompt frames \(\mathcal{F}_{vp}\), maintaining a cognitive anchor of the target instance appearance throughout reasoning; the latter endows the model with fine-grained local observation capability, enabling active filtering of keyframes and removal of redundant information when processing long videos with complex visual prompts. The two tools are formally defined as:
\begin{align}
    \mathcal{I}_{vp} &= \texttt{view\_visual\_prompt}\bigl(\mathcal{P}_{vp}\bigr), \quad \mathcal{P}_{vp} \in \mathbb{R}^{H \times W \times 3}, \label{eq:view_visual_prompt}\\
    \mathcal{V}_{crop} &= \texttt{crop\_video}\bigl(\mathcal{P}_v, \tau_s, \tau_e\bigr), \quad \tau_s, \tau_e \in \mathbb{R}^+, \ \tau_s < \tau_e,
\label{eq:crop_video}
\end{align}
where \(\mathcal{P}_{vp}\) denotes the visual prompt frame path and \(\mathcal{I}_{vp}\) represents the decoded image; \(\mathcal{P}_v\) denotes the video path, and \(\tau_s, \tau_e\) denote the start and end timestamps, respectively, yielding the cropped temporal segment \(\mathcal{V}_{crop}\). In each round \(t\) (where \(t = 0, 1, 2, \dots, T_{\max}\)), the model samples a response \(\mathcal{R}_t \sim \pi_{\theta}(\cdot \mid \mathcal{M})\) from the current message context \(\mathcal{M}\), which may contain \(\langle\text{tool\_call}\rangle\) blocks, \(\langle\text{answer}\rangle\) blocks, or both. When the model decides to invoke a perception tool, the tool is executed and its result is appended to \(\mathcal{M}\) for the next round; when an answer block appears, the \(\texttt{ExtractAnswer}\) function is called to extract answer \(A\), and the interaction terminates. This iterative cognitive cycle of ``active perception \(\rightarrow\) local zoom \(\rightarrow\) evidence-based reasoning'' parallels the human cognitive strategy of ``global browsing to local close-reading'' when confronting complex visual scenes, thereby circumventing the context loss and evidence obscuration inherent in single-pass compression paradigms.

To better illustrate the overall procedure, the entire rollout process is presented in Algorithm~\ref{alg:inference}.

\input{tables/algorithm_inference}

\subsection{Data Construction}
\label{sec:data_construction}

\textbf{Preliminary Data Curation.} To construct large-scale high-quality visual prompt video QA data, we propose a fully automated four-stage pipeline that transforms arbitrary video QA datasets into visual-prompt-dependent QA data without any manual annotation.
\begin{align}
\mathcal{D}_{final} = \mathcal{G}_4 \circ \mathcal{G}_3 \circ \mathcal{G}_2 \circ \mathcal{G}_1 (\mathcal{D}_{raw}),
\end{align}
where $\mathcal{G}_1$ to $\mathcal{G}_4$ correspond to Filtering, Verification, Mask Generation, and Rendering, respectively.

\textit{(1) Low-cost Text Filtering.} Since video tokens are computationally expensive, processing all data with video understanding leads to significant resource waste. We employ GPT-4o~\citep{openai2024gpt4o} to rapidly filter pure text QA pairs, eliminating samples unsuitable for visual prompting and preserving only QA pairs targeting concrete visual entities for the next stage:
\begin{equation}
\mathcal{F}_{filter}: \mathcal{D} \mapsto \{0,1\}, \quad
\mathcal{D}_{filter} = \{d \in \mathcal{D}_{raw} \mid \mathcal{F}_{filter}(d) = 1\},
\end{equation}
where $\mathcal{D}$ denotes the dataset space and $d = (v, q, a) \in \mathcal{D}$ contains video $v$, question $q$, and answer $a$.

\textit{(2) Video-level Verification.} For pre-filtered samples, we further verify whether the target is uniquely identifiable in the video. We use Gemini-3.1-Pro~\citep{gemini2025} to jointly process videos and original QA pairs through a five-step reasoning pipeline: target extraction with uniqueness judgment, generation of a unique semantic tag for SAM3 segmentation, temporal window localization, and QA rewriting with a unified $<vp>$ placeholder:
\begin{equation}
\mathcal{R}_{rewrite}: \mathcal{V} \times \mathcal{Q}\mathcal{A} \mapsto \mathcal{Q}\mathcal{A}_{vp}, \quad
\mathcal{Q}\mathcal{A}_{vp} = \mathcal{R}_{rewrite}(\mathcal{V}, \mathcal{Q}\mathcal{A}; \phi),
\end{equation}
where $\phi$ denotes the internal five-step reasoning process comprising target extraction with uniqueness judgment, semantic tag generation for SAM3, temporal window localization, and $<vp>$ substitution.

\textit{(3) Pixel-level Mask Generation.} Semantic tags alone are insufficient for pixel-level visual prompt rendering. We adopt SAM3~\citep{sam32025} to conduct text-driven video diffusion segmentation based on semantic tags, sampling at one frame per second to generate precise pixel-level masks:
\begin{equation}
\mathcal{M}_{\tau} = \text{SAM3}(\mathcal{V}, \tau; \omega), \quad \forall \tau \in \mathbb{T},
\quad \mathbb{T} = \left\{ \left\lfloor t \right\rfloor \mid t \in [0, T) \right\},
\end{equation}
where $\omega$ denotes the semantic tag condition and $T$ denotes the total video duration in seconds.

\textit{(4) Visual Prompt Rendering.} To enhance data diversity and establish alignment between visual prompt symbols and natural language descriptions, we uniformly sample eight visual prompt types and render them on video frames. We then invoke a language model to replace the $<vp>$ placeholder with natural language descriptions corresponding to the visual prompt types, producing visual prompt QA data ready for training:
\begin{equation}
\mathcal{Q}\mathcal{A}_{rendered} = \texttt{LLM}\bigl(\mathcal{Q}\mathcal{A}_{vp}, \mathcal{VP}\bigr),
\end{equation}
where $\mathcal{VP}$ denotes the sampled visual prompt type. The unified $<vp>$ facilitates community extensions by enabling seamless substitution across different visual prompt types without modifying downstream model interfaces.

\textbf{SFT and RL Data Curation.} Due to the limited capability of the base VLM, which exhibits poor instruction-following and high tool-calling error rates, we adopt a reject sampling strategy to generate high-quality multi-turn tool-calling trajectories. Specifically, we use data from the Preliminary Data Curation stage as input, and leverage Qwen3-VL-235B-A22B-Thinking to interact with the video environment using predefined tools. Subsequently, a rule-based discriminator filters out trajectories where the model responds correctly, ultimately yielding 34.2k high-quality samples for SFT stage. During the RL training phase, we further filter the SFT data based on the pass-k metric, resulting in 4.1k samples for GRPO training.

\subsection{Training Strategy}
\label{sec:training_strategy}

\textbf{Supervised Fine-Tuning.} We first conduct SFT to equip the model with foundational behaviors required for multimodal tool-calling VLMs, thereby ensuring effective interaction with the environment. Following the procedure described in Section~\ref{sec:data_construction}, we collect 34.2k high-quality trajectories for training. The model is trained by minimizing the standard autoregressive cross-entropy loss. The objective of SFT is to guide the model toward learning multi-turn, multi-scale active perception patterns in video environments, integrating visual evidence during reasoning, endowing the policy model with basic capabilities for interacting with the video environment, and establishing a foundation for agentic reinforcement learning.

\textbf{Agentic Reinforcement Learning.} In this stage, we treat the model as an agent capable of autonomously using tools, which actively decides whether to view the visual prompt, how to crop segments, and how to integrate retrieved evidence into the reasoning process. We employ GRPO to achieve this objective. The policy model is optimized by maximizing the following objective:

\input{tables/grpo}

where $r_{i,t}=\pi_\theta(y_{i,t}|x,y_{i,<t})/\pi_{\mathrm{old}}(y_{i,t}|x,y_{i,<t})$ and $\crr(r)=\clip(r,1-\epsilon,1+\epsilon)$. The rollout module samples a group of trajectories $\{y_1, y_2, \dots, y_G\}$ from the old policy $\pi_{\text{old}}$ for each input question $x$ through interaction with the external environment $\mathcal{V}$. The advantage term $\hat{A}_{i,t}$ is computed based on the relative rewards of outputs within each group. Additionally, we introduce a three-component reward modeling approach that jointly optimizes sampled trajectories across three dimensions: answer accuracy, format compliance, and generation efficiency. This design enhances final answer correctness, promotes more effective tool usage during inference, and produces more reliable and well-reasoned trajectories.

\noindent\textit{1. Answer Accuracy.} For the \(k\)-th rollout, let \(\hat{a}^{(k)}\) and \(a^\star\) denote the extracted answer and the ground truth, respectively. We adopt Qwen3-VL-235B-A22B-Instruct ~\citep{bai2025qwen3} as a judge to assess their semantic consistency and output a score in \(\{1, 0.5, 0\}\) (fully correct, partially correct, or incorrect). The accuracy reward is defined as:
\begin{equation}
R_{acc}^{(k)} = \operatorname{Judge}_{LLM}\!\big(\hat{a}^{(k)},\,a^\star\big) \;\in\; \{1,\; 0.5,\; 0\}.
\label{eq:reward_accuracy}
\end{equation}

\noindent\textit{2. Format Compliance.} Let \(y^{(k)}\) denote the complete textual output of the \(k\)-th rollout and \(\mathcal{S}\) be the predefined output schema. This reward encourages the model to consistently produce well-structured outputs with properly organized tool invocations and final answers, enabling reliable downstream parsing and verification. The format reward is computed as:
\begin{equation}
R_{format}^{(k)} = \mathbb{1}\!\big(y^{(k)} \text{ matches } \mathcal{S}\big).
\label{eq:reward_format}
\end{equation}

\noindent\textit{3. Parsimony Reward.} We introduce a parsimony reward to encourage the model to accomplish tasks with fewer tool-calling rounds while maintaining answer correctness. Specifically, let \(N^{(k)}\) denote the total number of perception tool invocations triggered in the \(k\)-th rollout. The parsimony reward is computed as:
\begin{equation}
R_{par}^{(k)} = \max\{0,\ 1 - \lambda \cdot N^{(k)}\},
\label{eq:reward_parsimony}
\end{equation}
where \(\lambda\) controls the strength of the parsimony penalty. This design implicitly incentivizes the model to only invoke tools when additional evidence is needed, thereby achieving a balance between effective reasoning and resource efficiency. 

\noindent\textit{4. Integrated Reward Function.} The final reward function is a weighted combination of the three
components described above, with weights used to balance the contributions of each component:
\begin{equation}
R^{(k)} = \alpha \cdot R_{acc}^{(k)} + \beta \cdot R_{format}^{(k)} + \gamma \cdot R_{par}^{(k)}.
\label{eq:reward_integrated}
\end{equation}
where \(\alpha + \beta + \gamma = 1\). By integrating these three components into the reward function, our VideoSeeker provides a comprehensive and fine-grained evaluation mechanism, guiding the model to better align with real-world application requirements when optimizing its reasoning capabilities.

%% file: tables/algorithm_inference.tex
\begin{algorithm}[h]
\small
\caption{Multi-turn Interactive Inference Process of VLM with Environment}
\label{alg:inference}
\begin{algorithmic}[1]
\REQUIRE Query $Q$, Visual Prompt Frame $\mathcal{F}_{vp}$, Video $\mathcal{V}$, Tool Set $\mathcal{T} = \{view\_visual\_prompt,\ crop\_video\}$, Policy Model $\pi_{\theta}$, Maximum Tool Rounds $T_{\max}$.
\ENSURE Final Answer $A$, Interaction Trajectory $\mathcal{Y}$, Tool Call History $\mathcal{H}$.

\STATE \textbf{Initialization:} $\mathcal{Y} \gets \emptyset$, $t \gets 0$, $\mathcal{H} \gets \emptyset$.
\STATE Encode $\mathcal{V}$ into visual frame sequence: $\mathcal{V}_{frames} \gets \texttt{EncodeVideoFrames}(\mathcal{V})$.
\STATE Compose user message $\mathcal{M} \gets \mathcal{V}_{frames} + \{Q,\ \texttt{ToolPrompt}(\mathcal{T},\ \mathcal{F}_{vp})\}$.

\WHILE{$t \leq T_{\max}$}
    \STATE Sample model response: $\mathcal{R}_t \sim \pi_{\theta}(\cdot \mid \mathcal{M})$.
    \STATE Append $\mathcal{R}_t$ to trajectory: $\mathcal{Y} \gets \mathcal{Y} + \mathcal{R}_t$.

    \IF{\texttt{<tool\_call><\//tool\_call>} detected in $\mathcal{R}_t$}
        \STATE Parse $\{(func_k,\ args_k)\}$ from $\mathcal{R}_t$, append to $\mathcal{H}$.
        \STATE Execute tools and append results to $\mathcal{M}$: $\mathcal{M} \gets \mathcal{M} + \texttt{ExecuteTools}(\{(func_k,\ args_k)\})$.
    \ENDIF

    \IF{\texttt{<answer><\//answer>} detected in $\mathcal{R}_t$}
        \STATE Extract answer $A \gets \texttt{ExtractAnswer}(\mathcal{R}_t)$.
        \STATE \textbf{return} $(A,\ \mathcal{Y},\ \mathcal{H})$.
    \ENDIF

    \STATE $t \gets t + 1$.
\ENDWHILE

\STATE \textbf{return} $(\texttt{NULL},\ \mathcal{Y},\ \mathcal{H})$.
\end{algorithmic}
\end{algorithm}

%% file: tables/grpo.tex
\begin{equation}
    \mathbb{E}_{x,\,\{y_i\}_{i=1}^G}\Bigg[
        \frac{1}{G}\sum_{i=1}^G\frac{1}{\sum_t I(y_{i,t})}
        \sum_{t:I(y_{i,t})=1}
        \min\!\big(r_{i,t},\;\crr(r_{i,t})\big)\,\hat{A}_{i,t}
    \Bigg]
    - \beta\,\KL(\pi_\theta\|\pi_{\mathrm{ref}}),
\label{eq:grpo}
\end{equation}

%% file: section/4_experiment.tex
\section{Experiments}
\label{sec:experiments}

\input{tables/bench_dim.tex}

\subsection{Implementation Details.}
\label{sec:impl_details}

\input{tables/bench_general.tex}

\textbf{Training and Evaluation Setup.}
In the SFT and RL stages, we leverage 34.2k trajectories and a curated dataset of 4.1k samples collected in Section~\ref{sec:data_construction}. All experiments are built upon Qwen3-VL-4B and Qwen3-VL-8B as base models. We evaluate VideoSeeker against a comprehensive suite of baselines, including open-source models like Video-R1~\citep{feng2025videor1}, VideoRFT~\citep{wang2025videorft}, Video-Thinker~\citep{liu2025videothinker} and proprietary models like GPT-4o~\citep{openai2024gpt4o}, Gemini-2.5-Pro~\citep{gemini2025}. Evaluations are conducted on four video understanding benchmarks: V2P-Bench~\citep{zhao2025v2pbench}, a dedicated instance-level video understanding evaluation framework, and three general video understanding benchmarks: Video-MME~\citep{li2024videomme}, LongVideoBench~\citep{fu2024longvideobench}, and LongVT~\citep{yang2025longvt}. We deploy models based on vLLM~\citep{kwon2023vllm} with native tool-calling mechanisms compatible with the OpenAI SDK, enabling multi-round tool-augmented reasoning. Specifically, we equip models with multiple visual tools, including frame sampling for temporal localization and object detection for spatial grounding, allowing models to dynamically invoke tools based on query complexity. For all evaluations, we set the temperature to 0 to ensure reproducibility of the results.

\textbf{Training Infrastructure.}
We conduct SFT on LLaMA-Factory~\citep{zheng2025llamafactory} and RL training on verl~\citep{zheng2025verl}, both employing full-parameter fine-tuning. All experiments are performed on 8 NVIDIA H800 GPUs. More detailed training hyperparameters are provided in Appendix~\ref{sec:appendix_hyperparams}.

\subsection{Main Results}

As illustrated in Table~\ref{tab:bench_dim}, our VideoSeeker series achieves the best performance among open-source models and is competitive with powerful closed-source models. Specifically, VideoSeeker-4B improves over the baseline Qwen3-VL-4B by +11.4\% on average, with particularly notable gains in HA, OD, and AS; scaling up to VideoSeeker-8B further improves over Qwen3-VL-8B by +13.7\% on average, showing clear advantages across most fine-grained dimensions while surpassing Gemini-2.5-Pro and GPT-4o. As shown in Table~\ref{tab:bench_general}, although our training data exclusively comes from instance-level video understanding tasks, VideoSeeker demonstrates generalization ability on general video understanding benchmarks, achieving an average improvement of +3.2\% and +3.3\% over three tasks. This indicates that our proposed tool-calling paradigm for instance-level video understanding can effectively transfer to broader general video understanding scenarios.

\subsection{Ablation Studies}

\begin{wraptable}{r}{0.3\textwidth}
    \small
    \vspace{-14pt}
    \caption{\textbf{Tools Ablation.}}
    \label{tab:tool_ablation}
    \resizebox{0.3\textwidth}{!}{
    \setlength{\tabcolsep}{6pt}
    \begin{tabular}{cc | c}
        \toprule
        \textbf{VP.} & \textbf{Crop.} & \textbf{Avg.} \\
        \midrule
        \multicolumn{2}{c|}{Qwen3-VL-8B \textit{(Baseline)}} & 60.8 \\
        \midrule
        \ding{51} &  & 69.4 \\
        & \ding{51} & 63.7 \\
        \ding{51} & \ding{51} & 74.5 \\
        \bottomrule
    \end{tabular}
    }
    \vspace{-13pt}
\end{wraptable}

\textbf{Tools Ablation.} As shown in Table~\ref{tab:tool_ablation}, we decomposed the tool set to analyze the contribution of each tool. The consistent performance improvements brought by the gradual introduction of the tool set clearly validate the effectiveness of our methodological paradigm. Notably, \textbf{the combination of the two tools yields synergistic gains} that exceed their individual contributions, indicating that the two tools form a complementary relationship in information acquisition.

\begin{wrapfigure}{l}{0.34\textwidth}
    \vspace{-14pt}
    \centering
    \includegraphics[width=\linewidth]{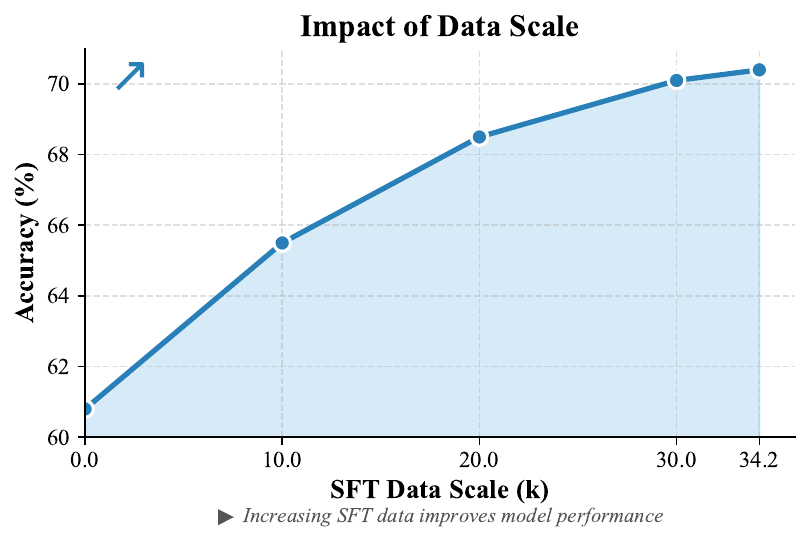}
    \caption{\textbf{Effect of Data Scale.}}
    \label{fig:data_scale}
    \vspace{-14pt}
\end{wrapfigure}

\textbf{Data Ablation.}
We construct several subsets by progressively increasing the sampling ratio from the full training corpus to investigate the impact of SFT data scale on model performance. As shown in Figure~\ref{fig:data_scale}, performance improves with increasing data volume, and the gains gradually diminish as the data scale expands further. This observation reveals a prominent \textbf{diminishing marginal returns pattern} in performance improvement, where the model approaches saturation beyond a certain data scale. These findings provide insights for balancing dataset scale and model performance.

\begin{wraptable}{r}{0.3\textwidth}
    \small
    \vspace{-12pt}
    \caption{\textbf{Reward Ablation.}}
    \label{tab:reward_ablation}
    \resizebox{0.3\textwidth}{!}{
    \setlength{\tabcolsep}{6pt}
    \begin{tabular}{ccc | c}
        \toprule
        \multicolumn{3}{c|}{\textbf{Reward Type}} & \multirow{2}*{\textbf{Acc.}} \\
        \textbf{$R_{acc}$} & \textbf{$R_{format}$} & \textbf{$R_{eff}$} & \\
        \midrule
        \ding{51} &  &  & 65.4 \\
        \ding{51} & \ding{51} &  & 73.1 \\
        \ding{51} &  & \ding{51} & 68.7 \\
        \ding{51} & \ding{51} & \ding{51} & 74.5 \\
        \bottomrule
    \end{tabular}
    }
    \vspace{-13pt}
\end{wraptable}

\textbf{Reward Ablation.} As shown in Table~\ref{tab:reward_ablation}, \textbf{our reward system provides a stable training signal}, and we systematically analyze the contribution of each reward signal during RL training. The format reward substantially outperforms the accuracy-only baseline, while the efficiency reward encourages more concise tool usage. Notably, the combined three-reward scheme surpasses the sum of individual contributions, revealing complementary effects across reward dimensions that jointly enhance effective reasoning.

\begin{wraptable}{l}{0.38\textwidth}
    \small
    \vspace{-14pt}
    \caption{\textbf{Stage Ablation.}}
    \label{tab:stage_ablation}
    \resizebox{0.38\textwidth}{!}{
    \setlength{\tabcolsep}{5pt}
    \begin{tabular}{cc c | c}
        \toprule
        \textbf{SFT} & \textbf{RL (Single Turn)} & \textbf{RL (Agentic)} & \textbf{Acc.} \\
        \midrule
        \multicolumn{3}{c}{Qwen3-VL-8B \textit{(Baseline)}} & 60.8 \\
        \midrule
        \ding{51} &  &  & 70.4 \\
        & \ding{51} &  & 62.6 \\
        &  & \ding{51} & 65.9 \\
        \ding{51} &  & \ding{51} & 74.5 \\
        \bottomrule
    \end{tabular}
    }
    \vspace{-8pt}
\end{wraptable}

\textbf{Stage Ablation.} As shown in Table~\ref{tab:stage_ablation}, we systematically ablate the contributions of the SFT and RL training stages to model performance. Experimental results demonstrate that high-quality SFT data endows the model with \textbf{robust reasoning patterns}, yielding a substantial performance boost (+9.6\%). In the zero-shot RL setting, single-turn RL leads to marginal improvement (+1.8\%). In contrast, agentic RL paradigm achieves +5.1\% improvement, which is more effective (+3.3\%) than single-turn RL. \textbf{This validates the agentic paradigm as a critical enabler for effective RL training} on instance-level video understanding tasks. The cascaded two-stage training paradigm leverages synergistic gains from both strategies, achieving optimal performance (74.5\%) and thereby establishing the optimal training pipeline in our framework.

\subsection{Analysis}

\textbf{Generalization to General Video Understanding Tasks.} Despite being trained exclusively on instance-level video understanding tasks, VideoSeeker demonstrates strong cross-task generalization on general video benchmarks (Table~\ref{tab:bench_general}), achieving +3.2\% and +3.3\% improvements in average. This reveals that core capabilities learned from instance-level tasks, such as long-range visual reasoning and multi-turn reasoning, transfer compositionally to broader video understanding scenarios. These findings highlight the value of instance-level video data in instilling generalizable priors, enabling cross-task improvements without additional general data.

\begin{wrapfigure}{l}{0.45\textwidth}
    \vspace{-10pt}
    \centering
    \includegraphics[width=\linewidth]{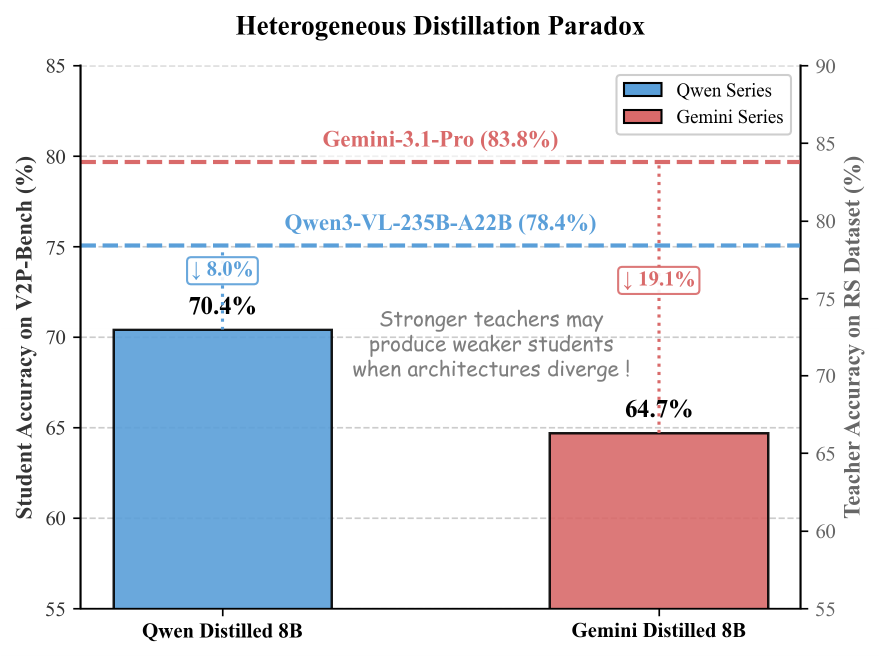}
    \vspace{-20pt}
    \caption{\textbf{Distillation Paradox.}}
    \label{fig:teacher_distillation_paradox}
    \vspace{-8pt}
\end{wrapfigure}

\textbf{The heterogeneous distillation paradox: stronger teachers may produce weaker students.} We experiment with two teacher models: Qwen3-VL-235B-A22B-Thinking and Gemini-3.1-Pro, achieving 78.4\% and 83.8\% accuracy on the rejection-sampled dataset respectively. After SFT training Qwen3-VL-8B, the resulting student models achieve 70.4\% and 64.7\% on V2P-Bench, with relative performance degradation of 8.0\% and 19.1\%. As illustrated in Figure~\ref{fig:teacher_distillation_paradox}, this reveals a counter-intuitive finding: \textbf{The raw capability of a teacher model does not proportionally transfer to distillation performance}. In homogeneous distillation, teachers and students share similar patterns, enabling efficient knowledge transfer; in heterogeneous distillation, pattern divergence is significant, causing stronger teachers' knowledge to be less effectively absorbed and leading to greater performance degradation.

\begin{wraptable}{l}{0.27\textwidth}
    \small
    \vspace{-14pt}
    \caption{\textbf{Reward Hacking.}}
    \label{tab:reward_hacking}
    \resizebox{0.27\textwidth}{!}{
    \setlength{\tabcolsep}{6pt}
    \begin{tabular}{cc | c}
        \toprule
        \textbf{MC} & \textbf{OE} & \textbf{Avg.} \\
        \midrule
        \multicolumn{2}{c|}{VideoSeeker-8B \textit{(SFT)}} & 70.4 \\
        \midrule
        \ding{51} &  & 43.8 \\
        & \ding{51} & 74.5 \\
        \bottomrule
    \end{tabular}
    }
    \vspace{-13pt}
\end{wraptable}

\textbf{RL training suffers from reward hacking on multiple-choice data.} As shown in Table~\ref{tab:reward_hacking}, RL training on multiple-choice (MC) data leads to a significant performance drop to 43.8\%, as models exploit random guessing rather than learning robust video understanding. In contrast, open-ended (OE) training achieves 74.5\%, demonstrating that OE with LLM judges provides a more robust strategy.

\begin{wrapfigure}{r}{0.34\textwidth}
    \vspace{-15pt}
    \centering
    \includegraphics[width=\linewidth]{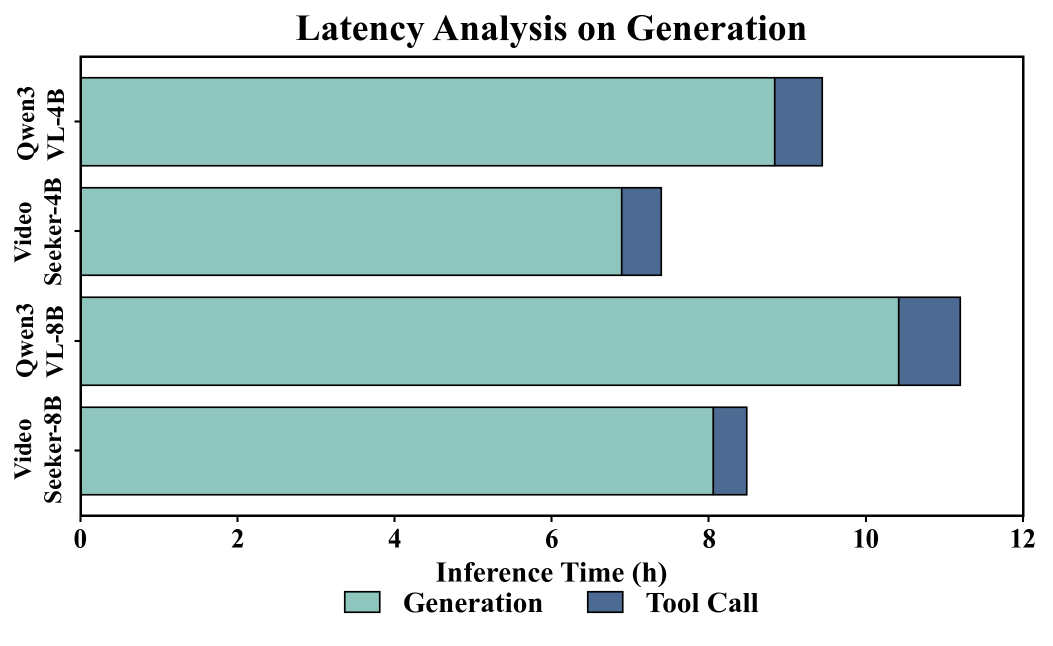}
    \vspace{-14pt}
    \caption{\textbf{Inference Latency.}}
    \label{fig:latency}
    \vspace{-15pt}
\end{wrapfigure}

\textbf{Time Efficiency.} We uniformly evaluate inference costs under the Agent mode. As illustrated in Figure~\ref{fig:latency}, VideoSeeker substantially reduces inference costs in both generation and tool-calling phases. Baseline models frequently exhibit frequent tool invocations accompanied by verbose chain-of-thought trajectories, resulting in prohibitively high computational overhead. In contrast, VideoSeeker converges to the correct answer with fewer total action steps through streamlined tool-calling strategies and more compact reasoning chains.

\textbf{Case Study.} The case study in Appendix \ref{sec:appendix_case_study} demonstrates how VideoSeeker successfully invokes tools to examine instance targets, clips specific segments for precise localization, and ultimately completes the task. This agentic interaction paradigm enables the model to handle instance-level video understanding with high precision, avoiding the localization errors inherent in traditional methods that rely on vague textual descriptions.

%% file: tables/bench_dim.tex
\begin{table}[t]
\centering
\caption{\textbf{Evaluation Results on V2P-Bench across Dimensions.} The "Agent" column indicates whether native tool calling is enabled (\ding{51}) or disabled (\ding{55}) in the prompt. The best results are \textbf{bold} and the second-best are \underline{underlined}.}
\resizebox{\textwidth}{!}{
\begin{tabular}{l | c |cccccccccccc c}

\toprule 
\textbf{Model} & \textbf{Agent} & \textbf{OA} & \textbf{HA} & \textbf{OD} & \textbf{FM} & \textbf{CR} & \textbf{PU} & \textbf{CI} & \textbf{FT} & \textbf{RT} & \textbf{AS} & \textbf{SR} & \textbf{GC} & \textbf{Avg.}\\
\midrule
\multicolumn{15}{c}{\textbf{Proprietary VLMs}} \\
\midrule
GPT-4o & \ding{55} & 76.6 & 68.9 & 41.3 & 60.8 & \underline{67.0} & \underline{73.3} & \textbf{67.6} & \textbf{68.1} & \underline{70.5} & 50.0 & 54.0 & 48.4 & 65.4 \\
Gemini-2.5-Pro & \ding{55} & 84.0 & 72.4 & \textbf{68.2} & \underline{71.8} & \textbf{75.0} & \underline{73.3} & 22.6 & \underline{66.7} & \textbf{72.7} & 47.4 & 67.5 & 63.6 & 69.8 \\
\midrule
\multicolumn{15}{c}{\textbf{Open-source VLMs}} \\
\midrule
LLaVA-OV-7B & \ding{55} & 57.1 & 52.1 & 28.3 & 47.1 & 63.8 & 59.1 & 41.0 & 42.1 & 35.6 & 63.2 & 62.8 & 43.2 & 52.8 \\
LLaVA-OV-72B & \ding{55} & 65.5 & 59.9 & 34.7 & 47.0 & 63.8 & 43.2 & 38.5 & 50.0 & 41.1 & 66.3 & 66.9 & 45.9 & 56.7 \\
InternVL3-8B & \ding{55} & 73.9 & 69.1 & 39.1 & 60.8 & 58.1 & 65.9 & 41.0 & 52.6 & 41.1 & 61.1 & 69.7 & 64.9 & 61.7 \\
LLaVA-Video-7B & \ding{55} & 60.5 & 58.1 & 37.0 & 49.0 & 62.9 & 54.5 & 41.0 & 52.6 & 48.9 & 57.9 & 56.6 & 40.5 & 54.8 \\
LLaVA-Video-72B & \ding{55} & 62.2 & 60.8 & 30.4 & 54.9 & 61.0 & 54.5 & 43.6 & 47.4 & 42.2 & 70.5 & 71.0 & 59.5 & 58.6 \\
\midrule
\multicolumn{15}{c}{\textbf{Ours}} \\
\midrule
Qwen3-VL-4B & \ding{55} & 73.7 & 67.1 & 41.2 & 58.8 & 51.7 & 52.9 & 51.2 & 42.1 & 42.2 & 63.2 & \underline{71.9} & 48.6 & 59.2 \\
Qwen3-VL-4B & \ding{51} & 73.0 & 68.6 & 26.7 & 52.9 & 26.9 & 62.5 & 41.0 & 50.0 & 44.5 & 67.6 & 64.8 & 40.5 & 57.6 \\
\rowcolor{softblue}VideoSeeker-4B & \ding{51} & \underline{85.7} & \textbf{79.4} & \underline{60.3} & 70.6 & 55.6 & 59.1 & 61.4 & 55.6 & 56.0 & \underline{76.5} & 69.2 & \underline{66.7} & \underline{70.6} \\
\color{softgreen}$\Delta$ & - & \color{softgreen}\textbf{+12.0} & \color{softgreen}\textbf{+12.3} & \color{softgreen}\textbf{+19.1} & \color{softgreen}\textbf{+11.8} & \color{softgreen}\textbf{+3.9} & \color{softgreen}\textbf{+6.2} & \color{softgreen}\textbf{+10.2} & \color{softgreen}\textbf{+13.5} & \color{softgreen}\textbf{+13.8} & \color{softgreen}\textbf{+13.3} & \color{softresult}\textbf{-2.7} & \color{softgreen}\textbf{+18.1} & \color{softgreen}\textbf{+11.4} \\
\midrule
Qwen3-VL-8B & \ding{55} & 81.8 & 68.1 & 41.3 & 67.8 & 45.4 & 58.3 & 58.3 & 47.4 & 38.5 & 56.2 & 70.5 & 51.3 & 60.8 \\
Qwen3-VL-8B & \ding{51} & 71.1 & 71.2 & 39.1 & 58.8 & 48.3 & 41.2 & 64.3 & 42.1 & 51.7 & 57.9 & 68.4 & 59.4 & 59.9 \\
\rowcolor{softblue}VideoSeeker-8B & \ding{51} & \textbf{94.3} & \underline{79.1} & 53.7 & \textbf{76.5} & 56.5 & \textbf{78.6} & \underline{66.8} & 62.3 & 55.0 & \textbf{78.8} & \textbf{78.8} & \textbf{70.3} & \textbf{74.5} \\
\color{softgreen}$\Delta$ & - & \color{softgreen}\textbf{+12.5} & \color{softgreen}\textbf{+11.0} & \color{softgreen}\textbf{+12.4} & \color{softgreen}\textbf{+8.7} & \color{softgreen}\textbf{+11.1} & \color{softgreen}\textbf{+20.3} & \color{softgreen}\textbf{+8.5} & \color{softgreen}\textbf{+14.9} & \color{softgreen}\textbf{+16.5} & \color{softgreen}\textbf{+22.6} & \color{softgreen}\textbf{+8.3} & \color{softgreen}\textbf{+19.0} & \color{softgreen}\textbf{+13.7} \\

\bottomrule 
\end{tabular}
}
\label{tab:bench_dim}
\end{table}

%% file: tables/bench_general.tex
\begin{wraptable}{r}{0.61\textwidth}
    \centering
    \vspace{-10pt}
    \caption{\textbf{Evaluation Results on General Benchmarks.} The bests are \textbf{bold} and the second-best are \underline{underlined}.}
    \vspace{-6pt}
    \label{tab:bench_general}
    \resizebox{0.99\linewidth}{!}
    {\small
    \begin{tabular}{l | c | c c c c}
    \toprule
    \textbf{Model} & \textbf{Agent} & \textbf{Video-MME} & \textbf{LongVideoBench} & \textbf{LongVT} & \textbf{Avg.} \\
    \midrule
    \multicolumn{5}{c}{\textbf{Proprietary VLMs}} \\
    \midrule
    GPT-4o & \ding{55} & \underline{77.2} & \textbf{81.3} & 17.4 & 58.6 \\
    Gemini-2.5 Pro & \ding{55} & \textbf{84.8} & - & - & - \\
    \midrule
    \multicolumn{5}{c}{\textbf{Open-source VLMs}} \\
    \midrule
    Video-R1-7B & \ding{55} & 61.0 & - & 27.9 & - \\
    VideoRFT-7B & \ding{55} & 60.9 & - & 26.5 & - \\
    Video-Thinker-7B & \ding{55} & 61.0 & - & 10.4 & - \\
    LongVT-7B & \ding{55} & 66.1 & - & 31.0 & - \\
    \midrule
    \multicolumn{5}{c}{\textbf{Ours}} \\
    \midrule
    Qwen3-VL-4B & \ding{55} & 65.3 & 62.6 & 38.5 & 55.5 \\
    Qwen3-VL-4B & \ding{51} & 61.5 & 50.4 & 36.9 & 49.6 \\
    \rowcolor{softblue}VideoSeeker-4B & \ding{51} & 66.1 & 64.2 & \underline{45.7} & \underline{58.7} \\
    \color{softgreen}$\Delta$ & - & \color{softgreen}\textbf{+0.8} & \color{softgreen}\textbf{+1.6} & \color{softgreen}\textbf{+7.2} & \color{softgreen}\textbf{+3.2} \\
    \midrule
    Qwen3-VL-8B & \ding{55} & 67.4 & 64.6 & 39.4 & 57.1 \\
    Qwen3-VL-8B & \ding{51} & 58.3 & 42.9 & 11.7 & 37.6 \\
    \rowcolor{softblue}VideoSeeker-8B & \ding{51} & 68.1 & \underline{66.5} & \textbf{46.5} & \textbf{60.4} \\
    \color{softgreen}$\Delta$ & - & \color{softgreen}\textbf{+0.7} & \color{softgreen}\textbf{+1.9} & \color{softgreen}\textbf{+7.1} & \color{softgreen}\textbf{+3.3} \\
    \bottomrule
    \end{tabular}
    }
\end{wraptable}

%% file: section/5_conclusion.tex
\section{Conclusion}
\label{sec:conclusion}

In this work, we propose VideoSeeker, an agentic paradigm that enables LVLMs to perform instance-level video understanding through native tool invocation. By integrating agentic reasoning with instance-level video understanding tasks, VideoSeeker empowers models to proactively perceive and retrieve relevant video segments on demand, achieving more precise spatial and temporal references than traditional text-only approaches. We construct a four-stage fully automated data synthesis pipeline to generate large-scale instance-level video data, and develop a two-stage training strategy to internalize tool-calling capabilities into LVLMs. Experiments on V2P-Bench demonstrate that VideoSeeker-8B achieves an average improvement of +13.7\%, surpassing GPT-4o and Gemini-2.5-Pro, while also exhibiting effective transferability to broader video understanding scenarios.

%% file: section/6_appendix.tex
\section*{Appendix Overview}
\label{sec:appendix}
~~$\bullet$ Section \ref{sec:appendix_dataset_details}: Dataset Details.

~~$\bullet$ Section \ref{sec:appendix_benchmark}: Benchmark Information.

~~$\bullet$ Section \ref{sec:appendix_hyperparams}: Hyperparameters.

~~$\bullet$ Section \ref{sec:appendix_limitations_social_impacts}: Limitations and Social Impacts.

~~$\bullet$ Section \ref{sec:appendix_training_curves}: Training Curves.

~~$\bullet$ Section \ref{sec:appendix_case_study}: Case Study.

~~$\bullet$ Section \ref{sec:appendix_prompts}: Prompts.

\section{Dataset Details}
\label{sec:appendix_dataset_details}

\begin{wraptable}{r}{0.4\textwidth}
    \centering
    \small
    \vspace{-14pt}
    \caption{\textbf{Video source distribution.}}
    \label{tab:video_source_distribution}
    \begin{tabular}{lcc}
        \toprule
        \textbf{Source} & \textbf{Count} & \textbf{Ratio} \\
        \midrule
        YouTube    & 5,405  & 72.1\% \\
        Charades   & 826    & 11.0\% \\
        ActivityNet & 526   & 7.0\%  \\
        YouCook2   & 454    & 6.1\%  \\
        NextQA     & 258    & 3.4\%  \\
        Ego4D      & 28     & 0.4\%  \\
        \bottomrule
    \end{tabular}
    \vspace{-13pt}
\end{wraptable}

\textbf{Data Source.} Our data construction pipeline uses the original videos and QA data from LLaVA-Video-178K~\citep{zhang2024llavavideo} as source material, comprising 178k videos and approximately 1.3 million instruction samples. This dataset integrates 10 mainstream video sources, covering domains such as activity recording, cooking, film, first-person perspective, and more. Multi-dimensional filtering rules are applied to select unedited raw videos with rich temporal variations, ensuring narrative completeness. Figure~\ref{tab:video_source_distribution} illustrates the source distribution of the video data.

\textbf{Data Construction Pipeline.} As described in Section~\ref{sec:data_construction}, we propose a fully automated four-stage pipeline, starting from 147,245 raw video QA samples from LLaVA-Video-178K~\citep{zhang2024llavavideo} and transforming them into visual-prompt-dependent QA data: \textit{(1) Text Filtering}, using GPT-4o to quickly pre-filter text QA and remove samples unsuitable for visual prompts (e.g., camera/cinematography questions, scene/background descriptions, overall activity summaries, counting questions, abstract/non-visual questions, ambient lighting/color queries), retaining 44.5\%; \textit{(2) Video Verification}, using Gemini-3.1-Pro to perform five-step reasoning jointly with the video, excluding multi-target ambiguous samples, retaining 32.9\%; \textit{(3) SAM3 Segmentation}, generating pixel-level masks at 1 fps based on semantic labels, retaining 27.9\%; \textit{(4) Visual Prompt Rendering}, uniformly sampling 8 visual prompt types (rectangle, mask contour, ellipse, triangle, scribble, point, arrow, set-of-mark) to render on video frames and rewrite QA, retaining 27.8\%. Table \ref{tab:pipeline_retention} presents detailed stage and statistics information. The final dataset covers 8 visual prompt types, providing diverse spatial and geometric variations for model training.

\begin{table}[htbp]
    \centering
    \small
    \caption{\textbf{Data pipeline retention statistics.}}
    \label{tab:pipeline_retention}
    \begin{tabular}{lcccl}
        \toprule
        \textbf{Stage} & \textbf{Output} & \textbf{Relative Retention} & \textbf{Description} \\
        \midrule
        Step 0: Original Dataset       & 147,245 & 100.0\% & LLaVA-Video-178K original data \\
        Step 1: Text Filtering         & 65,594  & 44.5\%  & GPT-4o filters text QA \\
        Step 2: Video Verification     & 48,457  & 32.9\%  & Gemini-3.1-Pro serves as a verifier \\
        Step 3: SAM3 Segmentation      & 41,083  & 27.9\%  & 1 FPS pixel-level mask generation \\
        Step 4: Visual Prompt Rendering & 40,929  & 27.8\%  & Rendering and QA rewriting \\
        \bottomrule
    \end{tabular}
\end{table}

\section{Benchmark Information}
\label{sec:appendix_benchmark}

We evaluate on four video understanding benchmarks, including one instance-level video understanding benchmark and three general video understanding benchmarks. This section introduces each benchmark. During evaluation, we uniformly segment videos into 256 frames on average.

\begin{itemize}
    \item \textbf{V2P-Bench~\citep{zhao2025v2pbench}} is a benchmark for evaluating LVLMs on visual-prompt-driven instance-level video understanding. Unlike text-only approaches, it introduces visual prompting to require precise spatial-temporal reasoning. It contains 980 videos with 1,172 QA pairs, covering three core tasks across twelve evaluation dimensions, assessing instance-level video understanding.

\item \textbf{Video-MME~\citep{li2024videomme}} is a video understanding benchmark for multimodal LLMs, evaluating capabilities in long-video and complex-reasoning scenarios. It contains approximately 900 manually curated videos spanning multiple domains, with 2.7K multiple-choice QA pairs. All data undergo rigorous human annotation. The dataset supports video, subtitles, and audio inputs. In our evaluation, we exclusively use video modality.

\item \textbf{LongVideoBench~\citep{fu2024longvideobench}} is a large-scale benchmark for long-context video-language understanding, evaluating multimodal models on videos up to one hour. It contains 3,763 diverse web videos covering movies, daily life, knowledge, and news, with 6,678 human-annotated multiple-choice questions. Video durations range from 8 seconds to 60 minutes. Its core innovation is the "Referring Reasoning" paradigm, embedding referring queries to locate relevant segments and requiring both precise retrieval and coherent contextual reasoning.

\item \textbf{LongVT~\citep{yang2025longvt}} is a benchmark for long-video open-domain question answering, containing 244 long videos and 1,280 QA pairs verified through manual review. The average video duration is approximately 1,688 seconds, with most videos (71.84\%) in the 15-30 minute range and 28.16\% exceeding 30 minutes. Its core design features a "needle-in-a-haystack" setting where supporting evidence exists only in narrow time windows, effectively evaluating models' abilities to locate and reason about fine-grained information within long videos.
\end{itemize}

\section{Hyperparameters}
\label{sec:appendix_hyperparams}
We detail the hyperparameters used in our training in Table~\ref{tab:SFT_hyperparameters} and Table~\ref{tab:RL_hyperparameters}. During agentic RL training, we set $\alpha = 0.8$, $\beta = 0.15$, $\gamma = 0.05$, with a sampling frame rate of 1, a maximum number of frames set to 256, and a maximum single-frame resolution set to 112896. During both SFT and RL training, LLaMA-Factory and verl automatically inject timestamps for videos, while during inference, we manually add corresponding timestamps to each frame.

\input{tables/hyperparameters.tex}

\section{Limitations and Social Impacts}
\label{sec:appendix_limitations_social_impacts}

While VideoSeeker demonstrates excellent performance on visual-prompt-driven video understanding tasks, it still has some limitations: First, our data construction pipeline relies on LLaVA-Video~\citep{zhang2024llavavideo} as the source, which means the generated data may inherit the domain bias and imbalance issues present in that dataset. On the positive side, VideoSeeker has the potential to enhance accessibility of video content, helping visually impaired users understand video content through intuitive visual prompts. However, similar to other vision models, the outputs may reflect biases in the training data, and we recommend thorough evaluation before applying it to critical scenarios.

\section{Training Curves}
\label{sec:appendix_training_curves}
\begin{figure}[htbp]
    \centering
    \includegraphics[width=0.9\textwidth]{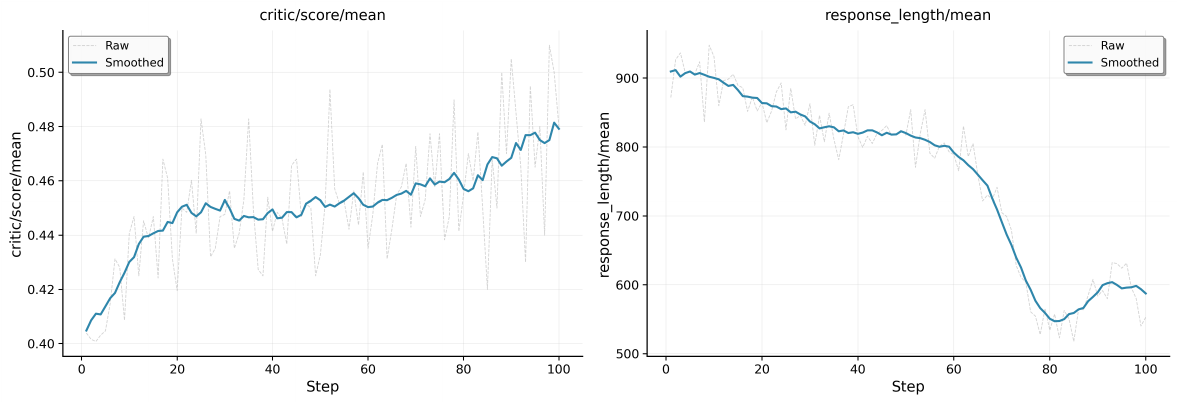}
    \caption{RL training curves.}
    \label{fig:training_curves}
\end{figure}

See Figure~\ref{fig:training_curves}.

\section{Case Study}
\label{sec:appendix_case_study}
\begin{figure}[htbp]
    \centering
    \includegraphics[width=0.9\textwidth]{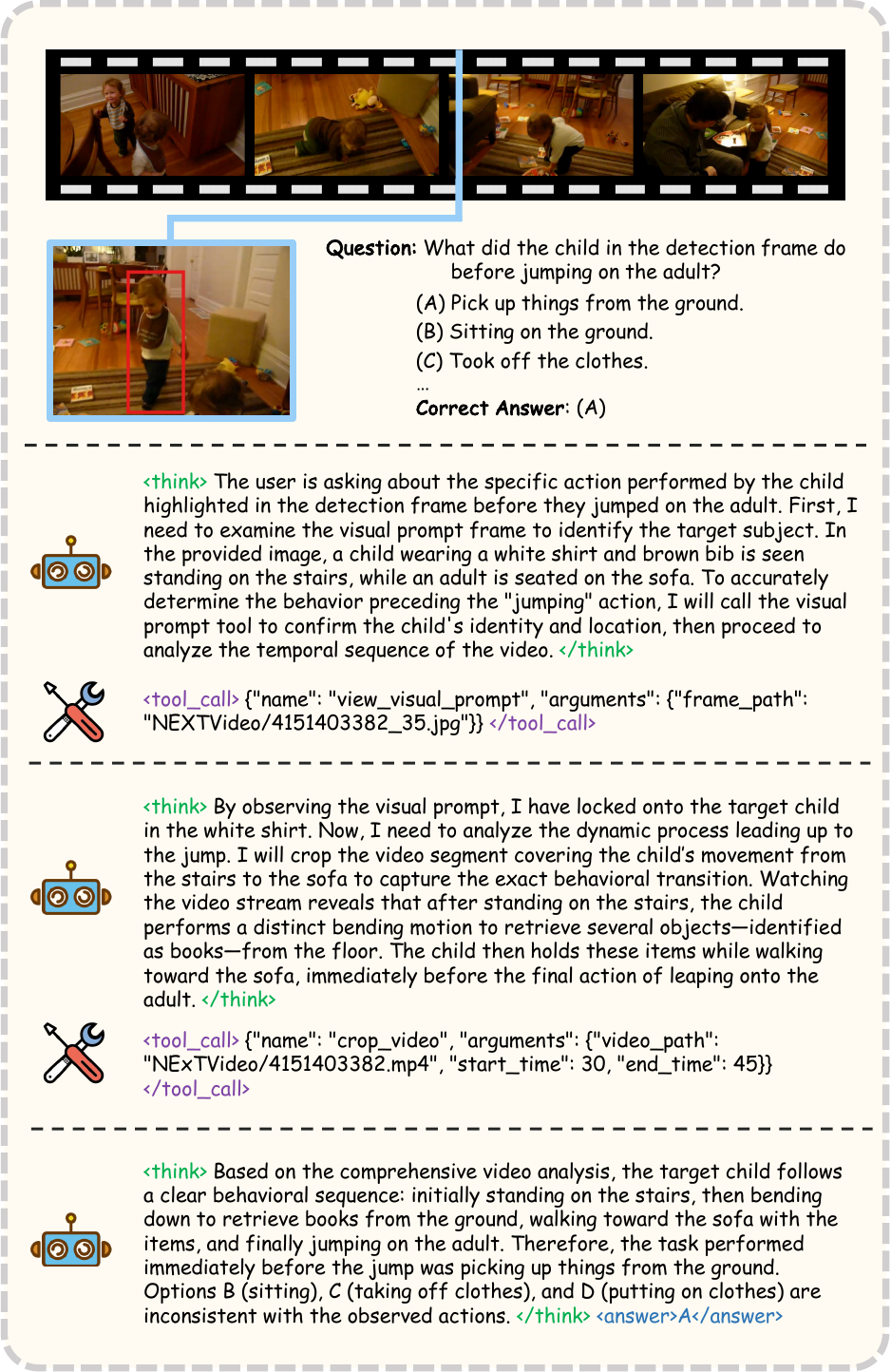}
    \caption{Case Study 1. The model invokes tools to proactively perceive instances and retrieve video segments, enabling instance-level video understanding tasks.}
    \label{fig:case1}
\end{figure}

\begin{figure}[htbp]
    \centering
    \includegraphics[width=0.9\textwidth]{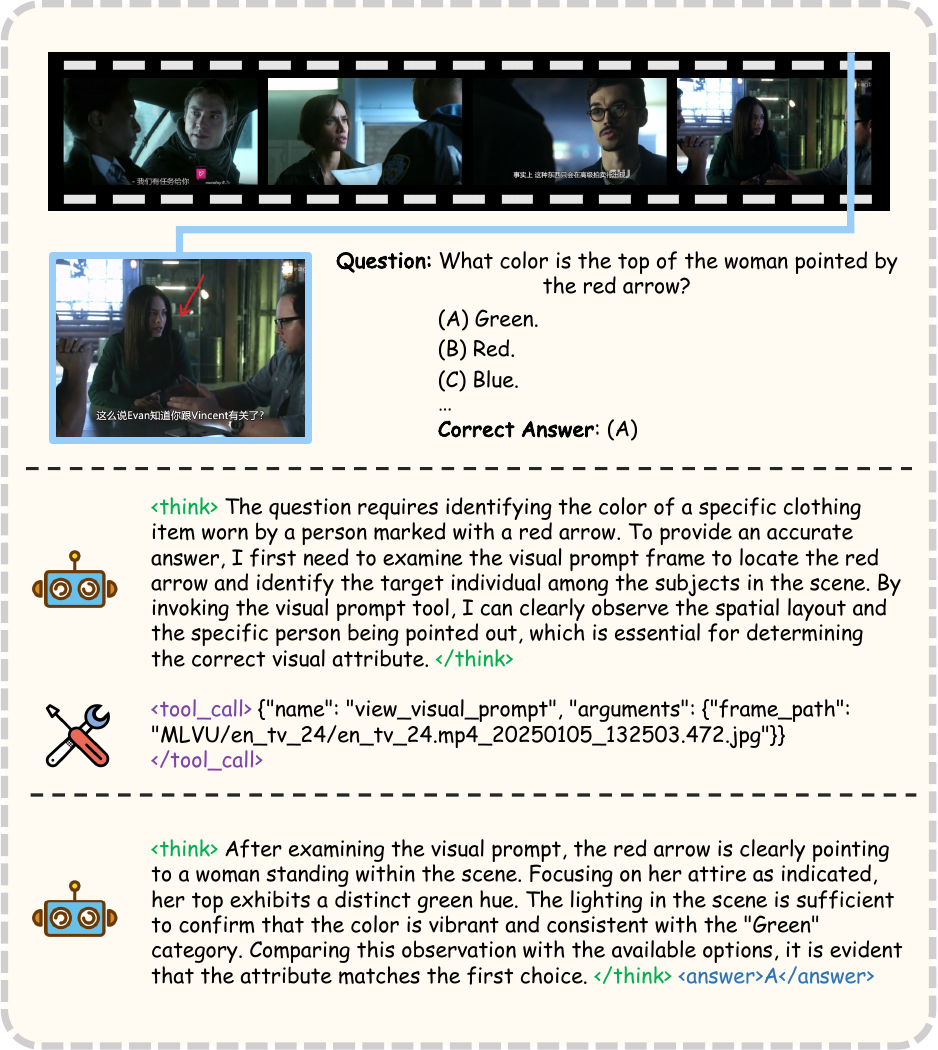}
    \caption{Case Study 2. The question only requires visual cue information, so the model adaptively invokes only the visual cue tool, avoiding unnecessary tool calls.}
    \label{fig:case2}
\end{figure}

See Figure~\ref{fig:case1} and Figure~\ref{fig:case2}.

\section{Prompts}
\label{sec:appendix_prompts}
In this section, we illustrate all the prompts used in our paper.

\subsection{Text Filtering Prompt}
\label{sec:prompt_text_filtering}
This prompt performs rapid pre-screening of QA samples to remove questions unsuitable for visual prompting (e.g., camera movements, scene backgrounds, counting). See Figure~\ref{prompt:text_filtering}.

\subsection{Video Verification Prompt}
\label{sec:prompt_video_verification}
This prompt guides five-step reasoning with the video: target extraction, uniqueness judgment, temporal localization, QA rewriting, and visual prompt type recommendation. See Figure~\ref{prompt:video_verify}.

\subsection{Rendering and Rewriting Prompt}
\label{sec:prompt_rendering_rewriting}
This prompt replaces target descriptions with generic visual prompt references, ensuring questions cannot be answered without visual prompting. See Figure~\ref{prompt:rendering_rewriting}.

\newpage

\input{appendix/filter_prompt.tex}

\input{appendix/video_verify.tex}

\input{appendix/rendering.tex}

%% file: tables/hyperparameters.tex
\begin{table}[h]
\centering
\begin{minipage}[t]{0.48\textwidth} 
\centering
\caption{Key hyperparameters for SFT.}
\label{tab:SFT_hyperparameters}
\begin{tabular}{@{}c|c@{}}
\toprule
\textbf{Name} & \textbf{Value} \\ \midrule
Finetuning type & Full \\
Freeze vision tower & True \\
Freeze multi-modal projector & False \\
Freeze language model & False \\
Cutoff len & 16384 \\
image/video max pixels & 112896 \\
Video FPS & 1.0 \\
Video max length & 256 \\
Batch size per device & 1 \\
Gradient accumulation steps & 2 \\
Learning rate & 1.0e-5 \\
LR scheduler type & cosine \\
Warmup ratio & 0.1 \\
Epochs & 1.0 \\
\bottomrule
\end{tabular}
\end{minipage}
\hfill 
\begin{minipage}[t]{0.48\textwidth} 
\centering
\caption{Key hyperparameters for RL.}
\label{tab:RL_hyperparameters}
\begin{tabular}{@{}c|c@{}}
\toprule
\textbf{Name} & \textbf{Value} \\ \midrule
Algorithm & GRPO \\
Max tool rounds & 5 \\
Agent loop & Tool agent \\
Rollout num & 8 \\
Train batch size & 32 \\
Mini batch size & 8 \\
Micro batch size per GPU & 1 \\
Learning rate & 1.0e-6 \\
KL loss coefficient & 0.001 \\
Entropy coefficient & 0.0 \\
Max prompt length & 16384 \\
Max response length & 4096 \\
Total epochs & 1 \\
GPU memory utilization & 0.8 \\
\bottomrule
\end{tabular}
\end{minipage}
\end{table}

%% file: appendix/filter_prompt.tex
\begin{tcolorbox}[breakable, colback=gray!5!white, colframe=gray!75!black,
    title=Text Filtering Prompt, boxrule=0.5mm, width=\textwidth, arc=3mm, auto outer arc,
    fontupper=\small, before skip=6pt]
    
    {\bfseries\large System Prompt:}\\[-6pt]
    
    \textbf{Task Definition: }
    
    You are a data filter for visual prompting QA construction. Given ONLY the text of a question-answer pair (no video), quickly determine whether the question is potentially suitable for a visual-prompted QA task — i.e., whether it targets a specific, concrete visual entity (person / object / animal) that could in principle be highlighted by a bounding box, arrow, or contour on a video frame.\\
    
    \textbf{Reject} (\texttt{is\_valid: false}) — clear disqualifiers
    
    Reject if the question clearly falls into ANY of the following types:
    \begin{itemize}
        \item \textbf{Camera / cinematography}: asks about camera movement, angle, zoom, transition, or how the video is shot.
        \textit{e.g., ``How does the camera move?'', ``What is the initial camera view?''}
    
        \item \textbf{Scene / background / setting}: asks about the overall scene, background, environment, setting, or atmosphere — not a specific foreground entity.
        \textit{e.g., ``What can be seen in the background?'', ``What is the primary setting of the video?''}
    
        \item \textbf{Overall / generic activity}: asks what the video is mainly about, without targeting a specific entity.
        \textit{e.g., ``What is the main activity in the video?'', ``What is the main focus?''}
    
        \item \textbf{Counting}: asks how many of something, rather than targeting one specific entity.
        \textit{e.g., ``How many shelves are visible?'', ``How many children are there?''}
    
        \item \textbf{Abstract / non-visual}: asks about abstract concepts, emotions, reasons without referencing a concrete visual entity; or targets non-visual properties (sound, smell, text labels, brand names on packaging).
        \textit{e.g., ``What indicates the cooking process?'', ``Which brand is on the jar?''}
    
        \item \textbf{Lighting / color of environment}: asks about ambient lighting changes or the color of background elements (walls, countertops) rather than a specific object.
    \end{itemize}
    
    \vspace{6pt}
    
    \textbf{Accept} (\texttt{is\_valid: true}):
    
    Accept if the question appears to target a specific, concrete visual entity:
    \begin{itemize}
        \item A person described by clothing, role, or action: \textit{``the person in the gray hoodie''}, \textit{``the worker with the helmet''}
        \item A specific named tool or object: \textit{``the wooden spoon''}, \textit{``the blue measuring cup''}, \textit{``the needle''}
        \item A specific animal: \textit{``the dog''}, \textit{``the kitten''}, \textit{``the calf''}
        \item A question asking WHERE / WHAT COLOR / WHAT ACTION / WHY about a clearly identifiable entity
    \end{itemize}
    When in doubt $\rightarrow$ accept (\texttt{is\_valid: true}). The downstream video analysis performs stricter uniqueness checking.
    
    Output ONLY valid JSON, no extra text.
    
    \textbf{Output Format}:\\
    
    Accept: \texttt{\{"is\_valid": true\}}\\
    Reject: \texttt{\{"is\_valid": false, "reason": "one short sentence"\}}
    
    \tcbline
    
    {\bfseries\large User Prompt:}\\[-6pt]
    
    Judge whether the following QA pair targets a specific visual entity suitable for visual prompting (text-only, no video).
    
    Here is the question: {\color{deepblue}\ttfamily \{question\}}\\
    Here is the options: {\color{deepblue}\ttfamily \{options\}}\\
    Here is the answer: {\color{deepblue}\ttfamily \{answer\}}\\
    
    Output JSON only.
    
    \end{tcolorbox}
    \captionof{figure}{Prompt of Text Filtering.}
    \vspace{10pt}
    \label{prompt:text_filtering}

%% file: appendix/video_verify.tex
\begin{tcolorbox}[breakable, colback=gray!5!white, colframe=gray!75!black,
    title=Video Verification Prompt, boxrule=0.5mm, width=\textwidth, arc=3mm, auto outer arc,
    fontupper=\small, before skip=6pt]

    {\bfseries\large System Prompt:}\\[-6pt]

    You are a video understanding expert specializing in visual prompting data construction.\\

    Your task: Given a video and an existing question-answer pair about the video, analyze whether the question targets a specific, uniquely identifiable object/person in the video, and if so, produce structured metadata for constructing visual-prompted QA data.\\

    \textbf{Definitions:}
    \smallskip
    
    - \textbf{Target}: The primary object or person that the question is asking about.
    \smallskip
    
    - \textbf{Uniqueness}: The target must be visually distinguishable from all other entities in the video. If multiple similar entities exist (e.g., ``a climber'' when there are several climbers), the target lacks uniqueness unless the description includes differentiating attributes (e.g., ``the climber in the red jacket'').
    \smallskip

    - \textbf{Visual Prompt}: A visual annotation (e.g., rectangle, contour, arrow, etc.) overlaid on a video frame to unambiguously indicate the target, replacing textual descriptions.
    \smallskip
    \smallskip

    \textbf{Workflow:}
    \smallskip

    \textbf{Step 1: Target Extraction \& Uniqueness Judgment}
    \smallskip

    Identify the target entity from the question. Determine:
    \begin{itemize}
        \item Is the question about a specific visual entity (person/object/animal)? Questions about abstract concepts, events, scenes, or the whole video are NOT target-specific.
        \item Is this target uniquely identifiable in the video? Watch the video and verify that no other entity could be confused with the target.
    \end{itemize}
    Output \texttt{is\_valid: false} if either condition fails, with a brief reason.\\
    If ambiguity exists, prefer rejecting (\texttt{is\_valid=false}) over guessing.
    \smallskip

    \textit{Examples:}
    \smallskip

    - \textit{``What colour shirt is the girl playing the violin wearing?''} $\rightarrow$ valid if exactly one such girl is identifiable.
    \smallskip

    - A tag like \textit{``climber scaling vertical rock face''} $\rightarrow$ invalid if multiple climbers appear without disambiguation.
    \smallskip
    \smallskip

    \textbf{Step 2: Generate Target Tag}
    \smallskip

    Produce a concise English tag (3-10 words) that:
    \begin{itemize}
        \item Uniquely identifies the target among all entities in the video
        \item Describes stable visual attributes (clothing, color, size, position, species, etc.)
        \item Is suitable as a text prompt for a detection/segmentation model (e.g., SAM3)
    \end{itemize}
    \smallskip
    \smallskip

    \textbf{Step 3: Temporal Localization}
    \smallskip

    Identify the primary continuous time window where the target is visible and relevant to the question:
    \begin{itemize}
        \item If the target appears in multiple segments, select the single most relevant segment
        \item Timestamps in seconds, rounded to 1 decimal place
    \end{itemize}
    \smallskip
    \smallskip

    \textbf{Step 4: QA Rewriting}
    \smallskip

    Rewrite the question by replacing the target's textual description with a visual prompt \texttt{<vp>}:
    \begin{itemize}
        \item Original: \textit{``What does the man playing the drums do with his feet?''}
        \item Rewritten: \texttt{``What does <vp> do with his feet?''}
    \end{itemize}
    For answer options: only rewrite options that explicitly mention the target; leave others unchanged.
    \smallskip
    \smallskip

    \textbf{Step 5: Visual Prompt Type Recommendation}
    \smallskip

    Select the most appropriate visual prompt type for this target from:
    \begin{itemize}
        \item \texttt{rectangle}: Bounding box. Best for well-bounded, non-overlapping targets.
        \item \texttt{mask\_contour}: Outline of segmentation mask. Best for irregularly shaped or partially occluded targets.
        \item \texttt{ellipse}: Elliptical highlight. Best for faces or compact targets.
        \item \texttt{triangle}: Triangular highlight around or pointing to the target.
        \item \texttt{scribble}: Freehand scribble over the target region.
        \item \texttt{point}: Single dot on target. Best for very small targets.
        \item \texttt{arrow}: Arrow pointing to target. Best for small or point-like targets.
        \item \texttt{set\_of\_mark}: Numeric label on target (e.g., ``the person marked with 1''). Best when multiple targets need simultaneous disambiguation.
    \end{itemize}
    Choose based on target size, shape, and scene complexity.\\

    \textbf{Output Format:}
    \smallskip

    Output ONLY valid JSON, no extra text.\\

    Here is a success sample:
    \begin{verbatim}
{
  "is_valid": true,
  "reason": null,
  "sample_id": "<copy from input if provided, else null>",
  "target_description": "the original textual description of the target",
  "tag": "concise unique English tag for detection/segmentation",
  "timestamp": {
    "start": 2.5,
    "end": 18.3
  },
  "rewritten_question": "question with target replaced by <vp>",
  "rewritten_options": ["A. option1", "B. option2", "..."],
  "rewritten_answer": "the correct answer (unchanged)",
  "visual_prompt_type": "rectangle"
}
    \end{verbatim}

    When \texttt{is\_valid} is false, only output:
    \begin{verbatim}
{
  "is_valid": false,
  "reason": "explanation of why this QA is not suitable"
}
    \end{verbatim}

    \tcbline

    {\bfseries\large User Prompt:}\\[-6pt]

    Here is the video: {\color{deepblue}\ttfamily \{video\}}

    Here is the question: {\color{deepblue}\ttfamily \{question\}}

    Here is the options: {\color{deepblue}\ttfamily \{options\}}

    Here is the answer: {\color{deepblue}\ttfamily \{answer\}}

    Analyze this QA pair following your workflow. Determine if the question targets a uniquely identifiable entity in the video, and if so, produce the complete structured output.

    \end{tcolorbox}
    \captionof{figure}{Prompt of Video Verification.}
    \vspace{10pt}
    \label{prompt:video_verify}

%% file: appendix/rendering.tex
\begin{tcolorbox}[breakable, colback=gray!5!white, colframe=gray!75!black,
    title=Rendering and Rewrite Prompt, boxrule=0.5mm, width=\textwidth, arc=3mm, auto outer arc,
    fontupper=\small, before skip=6pt]

    {\bfseries\large System Prompt:}\\[-6pt]

    You are an expert dataset writer for visual prompted video QA. Your job is to rewrite QA text so that answering requires the visual prompt on the frame, not the target's original textual attributes.
    \smallskip
    \smallskip

    \textbf{Context:}
    \smallskip

    - Upstream data already contains: \texttt{question}, \texttt{options}
    \smallskip

    - The placeholder \texttt{<vp>} marks where the target reference should be replaced by a visual prompt description.
    \smallskip
    \smallskip

    \textbf{Core objective:}
    \smallskip
    
    - Replace \texttt{<vp>} with a natural visual prompt phrase.
    \smallskip

    - Keep semantic meaning unchanged.
    \smallskip
    \smallskip

    \textbf{Hard constraints:}
    \smallskip

    - Do NOT change factual intent of the question.
    \smallskip

    - Do NOT alter option semantics, order, labels, or correctness.
    \smallskip

    - If an option does not contain \texttt{<vp>}, keep it exactly the same (except minimal grammar smoothing if absolutely necessary).
    \smallskip

    - Preserve language style and fluency.
    \smallskip

    - Never reveal concrete target identity in rewritten text.
    \smallskip
    \smallskip

    \textbf{Visual prompt phrase policy:}
    \smallskip

    Use generic, prompt-driven references:
    \begin{itemize}
        \item \texttt{rectangle} $\rightarrow$ ``the target in the highlighted box''
        \item \texttt{mask\_contour} $\rightarrow$ ``the target outlined by the contour''
        \item \texttt{ellipse} $\rightarrow$ ``the target inside the ellipse''
        \item \texttt{triangle} $\rightarrow$ ``the target indicated by the triangle marker''
        \item \texttt{scribble} $\rightarrow$ ``the target under the scribble mark''
        \item \texttt{point} $\rightarrow$ ``the target marked by the point''
        \item \texttt{arrow} $\rightarrow$ ``the target indicated by the arrow''
        \item \texttt{set\_of\_mark} $\rightarrow$ ``the target marked with number \{n\}'' (if number unknown, use ``the numbered target'')
    \end{itemize}
    \smallskip

    \textbf{Options handling:}
    \smallskip

    - Rewrite \texttt{rewritten\_answer} only when it contains \texttt{<vp>}.
    \smallskip

    - Otherwise keep it unchanged.
    \smallskip
    \smallskip

    \textbf{Output format:}
    \smallskip
    \begin{verbatim}
{
  "question_refined": [question],
}
    \end{verbatim}

    \tcbline

    {\bfseries\large User Prompt:}\\[-6pt]

    Please rewrite this sample for visual prompt dependency.
    \smallskip

    Here is the visual prompt type: {\color{deepblue}\ttfamily \{visual\_prompt\_type\}}

    Here is the question: {\color{deepblue}\ttfamily \{question\}}

    Here is the options: {\color{deepblue}\ttfamily \{options\}}

    Please rewrite the question and options according to the system instructions.

    \end{tcolorbox}
    \captionof{figure}{Prompt of Rendering and Rewriting.}
    \vspace{10pt}
    \label{prompt:rendering_rewriting}

%% file: section/99_check_list.tex
\section*{NeurIPS Paper Checklist}

\begin{enumerate}

\item {\bf Claims}
    \item[] Question: Do the main claims made in the abstract and introduction accurately reflect the paper's contributions and scope?
    \item[] Answer: \answerYes{} 
    \item[] Justification: The abstract and introduction clearly state our contributions.

\item {\bf Limitations}
    \item[] Question: Does the paper discuss the limitations of the work performed by the authors?
    \item[] Answer: \answerYes{} 
    \item[] Justification: We discuss limitations in Appendix~\ref{sec:appendix_limitations_social_impacts}.

\item {\bf Theory assumptions and proofs}
    \item[] Question: For each theoretical result, does the paper provide the full set of assumptions and a complete (and correct) proof?
    \item[] Answer: \answerNA{} 
    \item[] Justification: This paper does not present theoretical results or formal proofs

    \item {\bf Experimental result reproducibility}
    \item[] Question: Does the paper fully disclose all the information needed to reproduce the main experimental results of the paper to the extent that it affects the main claims and/or conclusions of the paper (regardless of whether the code and data are provided or not)?
    \item[] Answer: \answerYes{} 
    \item[] Justification: We fully disclose all key information required to reproduce the experimental results in the paper.

\item {\bf Open access to data and code}
    \item[] Question: Does the paper provide open access to the data and code, with sufficient instructions to faithfully reproduce the main experimental results, as described in supplemental material?
    \item[] Answer: \answerYes{} 
    \item[] Justification: The relevant datasets, code and models will be released publicly upon publication.

\item {\bf Experimental setting/details}
    \item[] Question: Does the paper specify all the training and test details (e.g., data splits, hyperparameters, how they were chosen, type of optimizer) necessary to understand the results?
    \item[] Answer: \answerYes{} 
    \item[] Justification: We provide complete details on all training and testing configurations, with detailed hyperparameters reported in Appendix~\ref{sec:appendix_hyperparams}.

\item {\bf Experiment statistical significance}
    \item[] Question: Does the paper report error bars suitably and correctly defined or other appropriate information about the statistical significance of the experiments?
    \item[] Answer: \answerNA{} 
    \item[] Justification: All evaluations use a consistent sampling temperature of 0.

\item {\bf Experiments compute resources}
    \item[] Question: For each experiment, does the paper provide sufficient information on the computer resources (type of compute workers, memory, time of execution) needed to reproduce the experiments?
    \item[] Answer: \answerYes{} 
    \item[] Justification: We report the computational resources in Section~\ref{sec:experiments}.
    
\item {\bf Code of ethics}
    \item[] Question: Does the research conducted in the paper conform, in every respect, with the NeurIPS Code of Ethics \url{https://neurips.cc/public/EthicsGuidelines}?
    \item[] Answer: \answerYes{} 
    \item[] Justification: The research conforms to the NeurIPS Code of Ethics.

\item {\bf Broader impacts}
    \item[] Question: Does the paper discuss both potential positive societal impacts and negative societal impacts of the work performed?
    \item[] Answer: \answerYes{} 
    \item[] Justification: We discuss potential positive and negative societal impacts in Section~\ref{sec:appendix_limitations_social_impacts}.
    
\item {\bf Safeguards}
    \item[] Question: Does the paper describe safeguards that have been put in place for responsible release of data or models that have a high risk for misuse (e.g., pre-trained language models, image generators, or scraped datasets)?
    \item[] Answer: \answerYes{} 
    \item[] Justification: VideoSeeker is an academic research project trained on publicly available datasets.

\item {\bf Licenses for existing assets}
    \item[] Question: Are the creators or original owners of assets (e.g., code, data, models), used in the paper, properly credited and are the license and terms of use explicitly mentioned and properly respected?
    \item[] Answer: \answerYes{} 
    \item[] Justification: We explicitly cite and comply with the licenses and usage terms of all datasets and assets used.

\item {\bf New assets}
    \item[] Question: Are new assets introduced in the paper well documented and is the documentation provided alongside the assets?
    \item[] Answer: \answerYes{} 
    \item[] Justification: All new assets are introduced and documented in the paper.

\item {\bf Crowdsourcing and research with human subjects}
    \item[] Question: For crowdsourcing experiments and research with human subjects, does the paper include the full text of instructions given to participants and screenshots, if applicable, as well as details about compensation (if any)? 
    \item[] Answer: \answerNA{} 
    \item[] Justification: VideoSeeker does not involve crowdsourcing nor research with human subjects.

\item {\bf Institutional review board (IRB) approvals or equivalent for research with human subjects}
    \item[] Question: Does the paper describe potential risks incurred by study participants, whether such risks were disclosed to the subjects, and whether Institutional Review Board (IRB) approvals (or an equivalent approval/review based on the requirements of your country or institution) were obtained?
    \item[] Answer: \answerNA{} 
    \item[] Justification: VideoSeeker does not involve research with human subjects.

\item {\bf Declaration of LLM usage}
    \item[] Question: Does the paper describe the usage of LLMs if it is an important, original, or non-standard component of the core methods in this research? Note that if the LLM is used only for writing, editing, or formatting purposes and does \emph{not} impact the core methodology, scientific rigor, or originality of the research, declaration is not required.
    \item[] Answer: \answerYes{} 
    \item[] Justification:  We only use LLMs for writing assistance, editing, and formatting, consistent with the NeurIPS policy on LLM usage.

\end{enumerate}

%% file: reference.bib
@article{qi2025vcrbench,
  title={Vcr-bench: A comprehensive evaluation framework for video chain-of-thought reasoning},
  author={Qi, Yukun and Zhao, Yiming and Zeng, Yu and Bao, Xikun and Huang, Wenxuan and Chen, Lin and Chen, Zehui and Zhao, Jie and Qi, Zhongang and Zhao, Feng},
  journal={arXiv preprint arXiv:2504.07956},
  year={2025}
}

@article{zeng2025jigsaw,
  title={Agentic Jigsaw Interaction Learning for Enhancing Visual Perception and Reasoning in Vision-Language Models},
  author={Zeng, Yu and Huang, Wenxuan and Huang, Shiting and Bao, Xikun and Qi, Yukun and Zhao, Yiming and Wang, Qiuchen and Chen, Lin and Chen, Zehui and Chen, Huaian and others},
  journal={arXiv preprint arXiv:2510.01304},
  year={2025}
}

@article{zhao2025v2pbench,
  title={V2p-bench: Evaluating video-language understanding with visual prompts for better human-model interaction},
  author={Zhao, Yiming and Zeng, Yu and Qi, Yukun and Liu, YaoYang and Bao, Xikun and Chen, Lin and Chen, Zehui and Miao, Qing and Liu, Chenxi and Zhao, Jie and others},
  journal={arXiv preprint arXiv:2503.17736},
  year={2025}
}

@inproceedings{zeng2025caption,
  title={Enhancing large vision-language models with ultra-detailed image caption generation},
  author={Zeng, Yu and Qi, Yukun and Zhao, Yiming and Bao, Xikun and Chen, Lin and Chen, Zehui and Huang, Shiting and Zhao, Jie and Zhao, Feng},
  booktitle={Proceedings of the 2025 Conference on Empirical Methods in Natural Language Processing},
  pages={26703--26729},
  year={2025}
}

@article{yang2025longvt,
  title={Longvt: Incentivizing" thinking with long videos" via native tool calling},
  author={Yang, Zuhao and Wang, Sudong and Zhang, Kaichen and Wu, Keming and Leng, Sicong and Zhang, Yifan and Li, Bo and Qin, Chengwei and Lu, Shijian and Li, Xingxuan and others},
  journal={arXiv preprint arXiv:2511.20785},
  year={2025}
}

@article{zheng2025deepeyes,
  title={Deepeyes: Incentivizing" thinking with images" via reinforcement learning},
  author={Zheng, Ziwei and Yang, Michael and Hong, Jack and Zhao, Chenxiao and Xu, Guohai and Yang, Le and Shen, Chao and Yu, Xing},
  journal={arXiv preprint arXiv:2505.14362},
  year={2025}
}

@article{zhang2025thinking,
  title={Thinking with videos: Multimodal tool-augmented reinforcement learning for long video reasoning},
  author={Zhang, Haoji and Gu, Xin and Li, Jiawen and Ma, Chixiang and Bai, Sule and Zhang, Chubin and Zhang, Bowen and Zhou, Zhichao and He, Dongliang and Tang, Yansong},
  journal={arXiv preprint arXiv:2508.04416},
  year={2025}
}

@article{su2025pixelreasoner,
  title={Pixel reasoner: Incentivizing pixel-space reasoning with curiosity-driven reinforcement learning},
  author={Wang, Haozhe and Su, Alex and Ren, Weiming and Lin, Fangzhen and Chen, Wenhu},
  journal={arXiv preprint arXiv:2505.15966},
  year={2025}
}

@article{wu2025vilasr,
  title={Reinforcing spatial reasoning in vision-language models with interwoven thinking and visual drawing},
  author={Wu, Junfei and Guan, Jian and Feng, Kaituo and Liu, Qiang and Wu, Shu and Wang, Liang and Wu, Wei and Tan, Tieniu},
  journal={arXiv preprint arXiv:2506.09965},
  year={2025}
}

@article{tian2025egor1,
  title={Ego-r1: Chain-of-tool-thought for ultra-long egocentric video reasoning},
  author={Tian, Shulin and Wang, Ruiqi and Guo, Hongming and Wu, Penghao and Dong, Yuhao and Wang, Xiuying and Yang, Jingkang and Zhang, Hao and Zhu, Hongyuan and Liu, Ziwei},
  journal={arXiv preprint arXiv:2506.13654},
  year={2025}
}

@article{zhao2025pyvision,
  title={Pyvision: Agentic vision with dynamic tooling},
  author={Zhao, Shitian and Zhang, Haoquan and Lin, Shaoheng and Li, Ming and Wu, Qilong and Zhang, Kaipeng and Wei, Chen},
  journal={arXiv preprint arXiv:2507.07998},
  year={2025}
}

@article{bai2025qwen3,
  title={Qwen3-vl technical report},
  author={Bai, Shuai and Cai, Yuxuan and Chen, Ruizhe and Chen, Keqin and Chen, Xionghui and Cheng, Zesen and Deng, Lianghao and Ding, Wei and Gao, Chang and Ge, Chunjiang and others},
  journal={arXiv preprint arXiv:2511.21631},
  year={2025}
}

@inproceedings{chen2024sharegpt4v,
  title={Sharegpt4v: Improving large multi-modal models with better captions},
  author={Chen, Lin and Li, Jinsong and Dong, Xiaoyi and Zhang, Pan and He, Conghui and Wang, Jiaqi and Zhao, Feng and Lin, Dahua},
  booktitle={European Conference on Computer Vision},
  pages={370--387},
  year={2024},
  organization={Springer}
}

@inproceedings{kwon2023vllm,
  title={Efficient memory management for large language model serving with pagedattention},
  author={Kwon, Woosuk and Li, Zhuohan and Zhuang, Siyuan and Sheng, Ying and Zheng, Lianmin and Yu, Cody Hao and Gonzalez, Joseph and Zhang, Hao and Stoica, Ion},
  booktitle={Proceedings of the 29th symposium on operating systems principles},
  pages={611--626},
  year={2023}
}

@article{zheng2025verl,
  title   = {HybridFlow: A Flexible and Efficient RLHF Framework},
  author  = {Guangming Sheng and Chi Zhang and Zilingfeng Ye and Xibin Wu and Wang Zhang and Ru Zhang and Yanghua Peng and Haibin Lin and Chuan Wu},
  year    = {2024},
  journal = {arXiv preprint arXiv: 2409.19256}
}

@inproceedings{zheng2025llamafactory,
  title={Llamafactory: Unified efficient fine-tuning of 100+ language models},
  author={Zheng, Yaowei and Zhang, Richong and Zhang, Junhao and Ye, Yanhan and Luo, Zheyan},
  booktitle={Proceedings of the 62nd annual meeting of the association for computational linguistics (volume 3: system demonstrations)},
  pages={400--410},
  year={2024}
}

@article{openai2024o1,
  title={Openai o1 system card},
  author={Jaech, Aaron and Kalai, Adam and Lerer, Adam and Richardson, Adam and El-Kishky, Ahmed and Low, Aiden and Helyar, Alec and Madry, Aleksander and Beutel, Alex and Carney, Alex and others},
  journal={arXiv preprint arXiv:2412.16720},
  year={2024}
}

@article{guo2025deepseekr1,
  title={Deepseek-r1: Incentivizing reasoning capability in llms via reinforcement learning},
  author={Guo, Daya and Yang, Dejian and Zhang, Haowei and Song, Junxiao and Wang, Peiyi and Zhu, Qihao and Xu, Runxin and Zhang, Ruoyu and Ma, Shirong and Bi, Xiao and others},
  journal={arXiv preprint arXiv:2501.12948},
  year={2025}
}

@article{rafailov2023direct,
  title={Direct preference optimization: Your language model is secretly a reward model},
  author={Rafailov, Rafael and Sharma, Archit and Mitchell, Eric and Manning, Christopher D and Ermon, Stefano and Finn, Chelsea},
  journal={Advances in neural information processing systems},
  volume={36},
  pages={53728--53741},
  year={2023}
}

@article{shao2024grpo,
  title={Deepseekmath: Pushing the limits of mathematical reasoning in open language models},
  author={Shao, Zhihong and Wang, Peiyi and Zhu, Qihao and Xu, Runxin and Song, Junxiao and Bi, Xiao and Zhang, Haowei and Zhang, Mingchuan and Li, YK and Wu, Yang and others},
  journal={arXiv preprint arXiv:2402.03300},
  year={2024}
}

@article{huang2025visionr1,
  title={Vision-r1: Incentivizing reasoning capability in multimodal large language models},
  author={Huang, Wenxuan and Jia, Bohan and Zhai, Zijie and Cao, Shaosheng and Ye, Zheyu and Zhao, Fei and Xu, Zhe and Tang, Xu and Hu, Yao and Lin, Shaohui},
  journal={arXiv preprint arXiv:2503.06749},
  year={2025}
}

@article{meng2025mmeureka,
  title={Mm-eureka: Exploring the frontiers of multimodal reasoning with rule-based reinforcement learning},
  author={Meng, Fanqing and Du, Lingxiao and Liu, Zongkai and Zhou, Zhixiang and Lu, Quanfeng and Fu, Daocheng and Han, Tiancheng and Shi, Botian and Wang, Wenhai and He, Junjun and others},
  journal={arXiv preprint arXiv:2503.07365},
  year={2025}
}

@article{yu2025perception,
  title={Perception-r1: Pioneering perception policy with reinforcement learning},
  author={Yu, En and Lin, Kangheng and Zhao, Liang and Yin, Jisheng and Wei, Yana and Peng, Yuang and Wei, Haoran and Sun, Jianjian and Han, Chunrui and Ge, Zheng and others},
  journal={arXiv preprint arXiv:2504.07954},
  year={2025}
}

@inproceedings{liu2025visualrft,
  title={Visual-rft: Visual reinforcement fine-tuning},
  author={Liu, Ziyu and Sun, Zeyi and Zang, Yuhang and Dong, Xiaoyi and Cao, Yuhang and Duan, Haodong and Lin, Dahua and Wang, Jiaqi},
  booktitle={Proceedings of the IEEE/CVF International Conference on Computer Vision},
  pages={2034--2044},
  year={2025}
}

@article{shen2025vlmr1,
  title={Vlm-r1: A stable and generalizable r1-style large vision-language model},
  author={Shen, Haozhan and Liu, Peng and Li, Jingcheng and Fang, Chunxin and Ma, Yibo and Liao, Jiajia and Shen, Qiaoli and Zhang, Zilun and Zhao, Kangjia and Zhang, Qianqian and others},
  journal={arXiv preprint arXiv:2504.07615},
  year={2025}
}

@inproceedings{yang2025r1,
  title={R1-onevision: Advancing generalized multimodal reasoning through cross-modal formalization},
  author={Yang, Yi and He, Xiaoxuan and Pan, Hongkun and Jiang, Xiyan and Deng, Yan and Yang, Xingtao and Lu, Haoyu and Yin, Dacheng and Rao, Fengyun and Zhu, Minfeng and others},
  booktitle={Proceedings of the IEEE/CVF International Conference on Computer Vision},
  pages={2376--2385},
  year={2025}
}

@article{feng2025videor1,
  title={Video-r1: Reinforcing video reasoning in mllms},
  author={Feng, Kaituo and Gong, Kaixiong and Li, Bohao and Guo, Zonghao and Wang, Yibing and Peng, Tianshuo and Wu, Junfei and Zhang, Xiaoying and Wang, Benyou and Yue, Xiangyu},
  journal={arXiv preprint arXiv:2503.21776},
  year={2025}
}

@article{li2025videochatr1,
  title={Videochat-r1: Enhancing spatio-temporal perception via reinforcement fine-tuning},
  author={Li, Xinhao and Yan, Ziang and Meng, Desen and Dong, Lu and Zeng, Xiangyu and He, Yinan and Wang, Yali and Qiao, Yu and Wang, Yi and Wang, Limin},
  journal={arXiv preprint arXiv:2504.06958},
  year={2025}
}

@article{wang2025videorft,
  title={Videorft: Incentivizing video reasoning capability in mllms via reinforced fine-tuning},
  author={Wang, Qi and Yu, Yanrui and Yuan, Ye and Mao, Rui and Zhou, Tianfei},
  journal={arXiv preprint arXiv:2505.12434},
  year={2025}
}

@article{wang2025timer1,
  title={Time-r1: Post-training large vision language model for temporal video grounding},
  author={Wang, Ye and Wang, Ziheng and Xu, Boshen and Du, Yang and Lin, Kejun and Xiao, Zihan and Yue, Zihao and Ju, Jianzhong and Zhang, Liang and Yang, Dingyi and others},
  journal={arXiv preprint arXiv:2503.13377},
  year={2025}
}

@article{gemini2025,
  title={Gemini 2.5: Pushing the frontier with advanced reasoning, multimodality, long context, and next generation agentic capabilities},
  author={Comanici, Gheorghe and Bieber, Eric and Schaekermann, Mike and Pasupat, Ice and Sachdeva, Noveen and Dhillon, Inderjit and Blistein, Marcel and Ram, Ori and Zhang, Dan and Rosen, Evan and others},
  journal={arXiv preprint arXiv:2507.06261},
  year={2025}
}

@article{sam32025,
  title={Sam 3: Segment anything with concepts},
  author={Carion, Nicolas and Gustafson, Laura and Hu, Yuan-Ting and Debnath, Shoubhik and Hu, Ronghang and Suris, Didac and Ryali, Chaitanya and Alwala, Kalyan Vasudev and Khedr, Haitham and Huang, Andrew and others},
  journal={arXiv preprint arXiv:2511.16719},
  year={2025}
}

@article{openai2024gpt4o,
  title={Gpt-4o system card},
  author={Hurst, Aaron and Lerer, Adam and Goucher, Adam P and Perelman, Adam and Ramesh, Aditya and Clark, Aidan and Ostrow, AJ and Welihinda, Akila and Hayes, Alan and Radford, Alec and others},
  journal={arXiv preprint arXiv:2410.21276},
  year={2024}
}

@article{fu2024longvideobench,
  title={Longvideobench: A benchmark for long-context interleaved video-language understanding},
  author={Wu, Haoning and Li, Dongxu and Chen, Bei and Li, Junnan},
  journal={Advances in Neural Information Processing Systems},
  volume={37},
  pages={28828--28857},
  year={2024}
}

@inproceedings{li2024videomme,
  title={Video-mme: The first-ever comprehensive evaluation benchmark of multi-modal llms in video analysis},
  author={Fu, Chaoyou and Dai, Yuhan and Luo, Yongdong and Li, Lei and Ren, Shuhuai and Zhang, Renrui and Wang, Zihan and Zhou, Chenyu and Shen, Yunhang and Zhang, Mengdan and others},
  booktitle={Proceedings of the IEEE/CVF conference on computer vision and pattern recognition},
  pages={24108--24118},
  year={2025}
}

@article{liu2025videothinker,
  title={Video-Thinker: Sparking" Thinking with Videos" via Reinforcement Learning},
  author={Wang, Shijian and Jin, Jiarui and Wang, Xingjian and Song, Linxin and Fu, Runhao and Wang, Hecheng and Ge, Zongyuan and Lu, Yuan and Cheng, Xuelian},
  journal={arXiv preprint arXiv:2510.23473},
  year={2025}
}

@article{chen2024mmstar,
  title={Are we on the right way for evaluating large vision-language models?},
  author={Chen, Lin and Li, Jinsong and Dong, Xiaoyi and Zhang, Pan and Zang, Yuhang and Chen, Zehui and Duan, Haodong and Wang, Jiaqi and Qiao, Yu and Lin, Dahua and others},
  journal={Advances in Neural Information Processing Systems},
  volume={37},
  pages={27056--27087},
  year={2024}
}

@article{zhang2024llavavideo,
  title={Llava-video: Video instruction tuning with synthetic data},
  author={Zhang, Yuanhan and Wu, Jinming and Li, Wei and Li, Bo and Ma, Zejun and Liu, Ziwei and Li, Chunyuan},
  journal={arXiv preprint arXiv:2410.02713},
  year={2024}
}

@misc{openai2025o3,
  title={Thinking with Images},
  author={OpenAI Team},
  year={2025},
  howpublished={\url{https://openai.com/index/thinking-with-images/}},
}

@article{chen2025scaling,
  title={Scaling rl to long videos},
  author={Chen, Yukang and Huang, Wei and Shi, Baifeng and Hu, Qinghao and Ye, Hanrong and Zhu, Ligeng and Liu, Zhijian and Molchanov, Pavlo and Kautz, Jan and Qi, Xiaojuan and others},
  journal={arXiv preprint arXiv:2507.07966},
  year={2025}
}

@article{deng2025openvlthinker,
  title={Openvlthinker: Complex vision-language reasoning via iterative sft-rl cycles},
  author={Deng, Yihe and Bansal, Hritik and Yin, Fan and Peng, Nanyun and Wang, Wei and Chang, Kai-Wei},
  journal={arXiv preprint arXiv:2503.17352},
  year={2025}
}

@article{wang2025adatooler,
  title={AdaTooler-V: Adaptive Tool-Use for Images and Videos},
  author={Wang, Chaoyang and Feng, Kaituo and Chen, Dongyang and Wang, Zhongyu and Li, Zhixun and Gao, Sicheng and Meng, Meng and Zhou, Xu and Zhang, Manyuan and Shang, Yuzhang and others},
  journal={arXiv preprint arXiv:2512.16918},
  year={2025}
}

@inproceedings{xu2025llavacot,
  title={Llava-cot: Let vision language models reason step-by-step},
  author={Xu, Guowei and Jin, Peng and Wu, Ziang and Li, Hao and Song, Yibing and Sun, Lichao and Yuan, Li},
  booktitle={Proceedings of the IEEE/CVF International Conference on Computer Vision},
  pages={2087--2098},
  year={2025}
}

@article{hong2025deepeyesv2,
  title={Deepeyesv2: Toward agentic multimodal model},
  author={Hong, Jack and Zhao, Chenxiao and Zhu, ChengLin and Lu, Weiheng and Xu, Guohai and Yu, Xing},
  journal={arXiv preprint arXiv:2511.05271},
  year={2025}
}

@inproceedings{deitke2025molmo,
  title={Molmo and pixmo: Open weights and open data for state-of-the-art vision-language models},
  author={Deitke, Matt and Clark, Christopher and Lee, Sangho and Tripathi, Rohun and Yang, Yue and Park, Jae Sung and Salehi, Mohammadreza and Muennighoff, Niklas and Lo, Kyle and Soldaini, Luca and others},
  booktitle={Proceedings of the Computer Vision and Pattern Recognition Conference},
  pages={91--104},
  year={2025}
}

@article{xing2025caprl,
  title={Caprl: Stimulating dense image caption capabilities via reinforcement learning},
  author={Xing, Long and Dong, Xiaoyi and Zang, Yuhang and Cao, Yuhang and Liang, Jianze and Huang, Qidong and Wang, Jiaqi and Wu, Feng and Lin, Dahua},
  journal={arXiv preprint arXiv:2509.22647},
  year={2025}
}

@article{hong2026glm5vturbo,
  title={GLM-5V-Turbo: Toward a Native Foundation Model for Multimodal Agents},
  author={Hong, Wenyi and Gu, Xiaotao and Pan, Ziyang and Yang, Zhen and Wang, Yuting and Wang, Yue and Yue, Yuanchang and Wang, Yu and Wang, Yanling and Wang, Yan and others},
  journal={arXiv preprint arXiv:2604.26752},
  year={2026}
}

@article{wang2025internvl3p5,
  title={Internvl3. 5: Advancing open-source multimodal models in versatility, reasoning, and efficiency},
  author={Wang, Weiyun and Gao, Zhangwei and Gu, Lixin and Pu, Hengjun and Cui, Long and Wei, Xingguang and Liu, Zhaoyang and Jing, Linglin and Ye, Shenglong and Shao, Jie and others},
  journal={arXiv preprint arXiv:2508.18265},
  year={2025}
}

@article{clark2026molmo2,
  title={Molmo2: Open Weights and Data for Vision-Language Models with Video Understanding and Grounding},
  author={Clark, Christopher and Zhang, Jieyu and Ma, Zixian and Park, Jae Sung and Salehi, Mohammadreza and Tripathi, Rohun and Lee, Sangho and Ren, Zhongzheng and Kim, Chris Dongjoo and Yang, Yinuo and others},
  journal={arXiv preprint arXiv:2601.10611},
  year={2026}
}

@article{team2026kimik2p5,
  title={Kimi K2. 5: Visual Agentic Intelligence},
  author={Team, Kimi and Bai, Tongtong and Bai, Yifan and Bao, Yiping and Cai, SH and Cao, Yuan and Charles, Y and Che, HS and Chen, Cheng and Chen, Guanduo and others},
  journal={arXiv preprint arXiv:2602.02276},
  year={2026}
}

@article{chen2025minimax,
  title={Minimax-m1: Scaling test-time compute efficiently with lightning attention},
  author={Chen, Aili and Li, Aonian and Gong, Bangwei and Jiang, Binyang and Fei, Bo and Yang, Bo and Shan, Boji and Yu, Changqing and Wang, Chao and Zhu, Cheng and others},
  journal={arXiv preprint arXiv:2506.13585},
  year={2025}
}

@inproceedings{ren2024timechat,
  title={Timechat: A time-sensitive multimodal large language model for long video understanding},
  author={Ren, Shuhuai and Yao, Linli and Li, Shicheng and Sun, Xu and Hou, Lu},
  booktitle={Proceedings of the IEEE/CVF Conference on Computer Vision and Pattern Recognition},
  pages={14313--14323},
  year={2024}
}
